\PassOptionsToPackage{unicode}{hyperref}
\PassOptionsToPackage{hyphens}{url}
\PassOptionsToPackage{dvipsnames,svgnames,x11names}{xcolor}
\documentclass[
  10pt,
  letterpaper,
  10pt]{article}
\usepackage{xcolor}
\usepackage[margin=1in]{geometry}
\usepackage{amsmath,amssymb}
\usepackage{iftex}
\ifPDFTeX
  \usepackage[T1]{fontenc}
  \usepackage[utf8]{inputenc}
  \usepackage{textcomp} 
\else 
  \usepackage{unicode-math} 
  \defaultfontfeatures{Scale=MatchLowercase}
  \defaultfontfeatures[\rmfamily]{Ligatures=TeX,Scale=1}
\fi
\usepackage{lmodern}
\ifPDFTeX\else
\fi
\IfFileExists{upquote.sty}{\usepackage{upquote}}{}
\IfFileExists{microtype.sty}{
  \usepackage[]{microtype}
  \UseMicrotypeSet[protrusion]{basicmath} 
}{}
\usepackage{setspace}
\makeatletter
\@ifundefined{KOMAClassName}{
  \IfFileExists{parskip.sty}{%
    \usepackage{parskip}
  }{
    \setlength{\parindent}{0pt}
    \setlength{\parskip}{6pt plus 2pt minus 1pt}}
}{
  \KOMAoptions{parskip=half}}
\makeatother
\usepackage{longtable,booktabs,array}
\usepackage{caption}
\usepackage{calc} 
\usepackage{etoolbox}
\makeatletter
\patchcmd\longtable{\par}{\if@noskipsec\mbox{}\fi\par}{}{}
\makeatother
\IfFileExists{footnotehyper.sty}{\usepackage{footnotehyper}}{\usepackage{footnote}}
\makesavenoteenv{longtable}
\usepackage{graphicx}
\makeatletter
\newsavebox\pandoc@box
\newcommand*\pandocbounded[1]{
  \sbox\pandoc@box{#1}%
  \Gscale@div\@tempa{\textheight}{\dimexpr\ht\pandoc@box+\dp\pandoc@box\relax}%
  \Gscale@div\@tempb{\linewidth}{\wd\pandoc@box}%
  \ifdim\@tempb\p@<\@tempa\p@\let\@tempa\@tempb\fi
  \ifdim\@tempa\p@<\p@\scalebox{\@tempa}{\usebox\pandoc@box}%
  \else\usebox{\pandoc@box}%
  \fi%
}
\def\fps@figure{htbp}
\makeatother
\NewDocumentCommand\citeproctext{}{}
\NewDocumentCommand\citeproc{mm}{%
  \begingroup\def\citeproctext{#2}\cite{#1}\endgroup}
\makeatletter
 \let\@cite@ofmt\@firstofone
 \def\@biblabel#1{}
 \def\@cite#1#2{{#1\if@tempswa , #2\fi}}
\makeatother
\newlength{\cslhangindent}
\newlength{\csllabelwidth}
\newenvironment{CSLReferences}[2] 
 {\begin{list}{}{%
  \setlength{\itemindent}{0pt}
  \setlength{\leftmargin}{0pt}
  \setlength{\parsep}{0pt}
  \ifodd #1
   \setlength{\leftmargin}{\cslhangindent}
   \setlength{\itemindent}{-1\cslhangindent}
  \fi
  \setlength{\itemsep}{#2\baselineskip}}}
 {\end{list}}
\usepackage{calc}

\providecommand{\tightlist}{%
  \setlength{\itemsep}{0pt}\setlength{\parskip}{0pt}}
\usepackage{booktabs}
\usepackage{hyperref}
\IfFileExists{orcidlink.sty}{\usepackage{orcidlink}}{\newcommand{\orcidlink}[1]{}}
\usepackage{longtable}
\usepackage{array}
\usepackage{microtype}
\usepackage{amsmath,amssymb}
\usepackage{graphicx}
\usepackage{xcolor}
\usepackage{float}
\usepackage{enumitem}
\setlist{nosep}
\usepackage{bookmark}
\IfFileExists{xurl.sty}{\usepackage{xurl}}{} 
\makeatletter
\@ifundefined{xmpquote}{}{}
\makeatother
\hypersetup{
  pdftitle={Reproducible Evaluation of MoE Expert Caching},
  pdfauthor={Yu Zhang  China National Chemical Equipment Co.~Ltd. ORCID: },
  colorlinks=true,
  linkcolor={blue},
  filecolor={Maroon},
  citecolor={blue},
  urlcolor={blue},
  pdfcreator={LaTeX via pandoc}}

\title{Reproducible Evaluation of MoE Expert Caching}
\usepackage{etoolbox}
\makeatletter
\providecommand{\subtitle}[1]{
  \apptocmd{\@title}{\par {\large #1 \par}}{}{}
}
\makeatother
\subtitle{Replay Semantics, Workload Contamination, and Operating
Regimes}
\author{Yu Zhang \orcidlink{0009-0000-8884-6497}\\
China National Chemical Equipment Co.~Ltd.\\
ORCID:
\href{https://orcid.org/0009-0000-8884-6497}{0009-0000-8884-6497}}
\date{2026-08-04}

\begin{document}
\maketitle

\setstretch{1}
\begin{center}\rule{0.5\linewidth}{0.5pt}\end{center}

\subsection{Abstract}\label{abstract}

Mixture-of-Experts (MoE) models have outgrown the high-bandwidth memory
of the accelerators that serve them, and offloading expert weights to
host memory has become a standard response. This makes expert cache
management an attractive lever: if a smarter policy raised the hit rate,
the same model would need less expert traffic per token. Evaluating that
hypothesis is a measurement problem, and we find the measurement
fragile.

Using a trace-driven, event-atomic simulator over three MoE models (40,
64 and 128 experts), we isolate three evaluation axes that change
conclusions rather than shift numbers. \emph{Replay semantics}: under a
fused-event traffic contract, an inconsistent per-access replay inflates
recency-based policies by 27--29\% while leaving frequency-based and
static policies within 4\%, inverting the policy ranking. \emph{Workload
contamination}: probe sets using one instruction template per category
produce verbatim-identical generation prefixes; a matched-pair rendering
intervention moves the measured early-window effect by 19.4--31.9
percentage points and reverses which workloads appear most
cache-friendly. \emph{Operating regimes}: normalized miss fractions do
not transfer across models, so the per-step expert union relative to
per-layer capacity must be reported --- though permuting only the
temporal order of an otherwise identical event stream moves the
offline-optimal gap from 44.9\% to 30.8\%, so it is not sufficient.

After correcting all three, a stable gap to the offline optimum remains
(44.2--45.9\% across 13 frozen workload compositions). A
forced-admission oracle attributes 84.3--96.6\% of it to knowing which
resident expert is used furthest in the future. A causal next-use
predictor, used as an eviction rule, recovers \(-11.4\)\% of the gap; at
the decision points it selects an optimal victim 3.4\% of the time
against 2.4\% for a random resident block and 20.6-22.1\% for LRU and
LFRU. Our position is narrow: \textbf{in our evaluated settings a large
offline-optimal gap substantially overstates the gains recovered by
representative lightweight causal mechanisms.} On publication we will
release the simulator, a diversity-controlled probe set, the
contamination diagnostics and a reporting checklist, subject to the
upstream licences recorded in the artifact matrix.

\begin{center}\rule{0.5\linewidth}{0.5pt}\end{center}

\subsection{1. Introduction}\label{introduction}

Serving a frontier Mixture-of-Experts model no longer fits the memory it
was designed for. A contemporary large MoE dedicates on the order of 1.3
TiB to routed expert weights while activating only a small fraction per
token, so recent systems keep a subset of experts resident in device
memory and stream the rest from host memory on demand. Variants appear
in open-source prototypes, research systems, and inference-engine
proposals; their execution contracts and performance targets differ
(\citeproc{ref-shazeer2017sparselygated}{Shazeer et al. 2017};
\citeproc{ref-fedus2022switch}{Fedus et al. 2022};
\citeproc{ref-eliseev2023mixtraloffloading}{Eliseev and Mazur 2023};
\citeproc{ref-xue2025moeinfinity}{Xue et al. 2025};
\citeproc{ref-tang2024hobbit}{Tang et al. 2024}).

The design also creates an obvious hypothesis. If experts are cached,
then a better cache policy should raise the hit rate, lower
host-to-device traffic, and let the same model run on fewer
accelerators. A workload-aware controller --- one that learns which
experts a customer's traffic actually touches --- would then be worth
building. A substantial body of recent work pursues variants of this
idea through prefetching, learned routing prediction, and specialized
replacement policies (\citeproc{ref-du2024sidamoe}{Du et al. 2024};
\citeproc{ref-fang2025fate}{Fang et al. 2025};
\citeproc{ref-hoang2026specmd}{Hoang et al. 2026};
\citeproc{ref-zhu2026dali}{Zhu et al. 2026}).

Whether the hypothesis holds is an empirical question, and answering it
requires measuring how much room a policy has: the distance between the
best causal policy and an offline optimum. We set out to measure exactly
that. What we found first was that the measurement itself is unstable in
ways that are easy to miss and that change the answer, not just its
precision.

\textbf{Three axes.} Within our trace-driven study, we identify three
evaluation choices, each individually plausible and each capable of
reversing a conclusion. Their applicability is not identical: Axis I
concerns replay of a claimed fused-event contract and does not apply
retrospectively to a real end-to-end system whose implementation defines
its own transfer semantics (§12); Axes II and III concern workload
construction and operating-regime reporting more broadly.

\emph{Replay semantics.} Under the fused-event traffic contract studied
here, the experts a scheduling step touches at a layer form one
committed execution unit. A simulator that flattens this unit into
individual accesses can evict, part-way through the event, an expert
that was resident at event start and is required later, then count an
artificial refetch. The distortion is not uniform noise. On
Qwen3-30B-A3B at \ensuremath{\rho}=40\% and B=8, sequential replay
inflates LRU by 27.4\%, LFRU by 28.5\%, and Least-Stale by 27.4\%, but
LFU by only 3.4\%, Belady by 4.3\%, and a same-trace static diagnostic
not at all. Because it penalizes exactly one family of policies, it
inverts their ranking: under sequential replay the same-trace static
diagnostic appears to beat every dynamic policy; under event-atomic
replay LFRU beats it by 5.7\%. This is a replay-invariance comparison,
not a deployable static baseline (§2.3). In our own earlier
measurements, correcting this semantics moved the steady-state prefill
gap on two models from a range that looked like algorithmic headroom
down to 0.155--3.169\%.

\emph{Workload contamination.} Studies of workload-conditioned routing
need probe sets partitioned by task. The natural construction --- one
instruction template per category, filled with different payloads ---
turns out to be hazardous. Under raw continuation, or within a short
decode window, the model reproduces the shared template before it
produces anything task-specific, so concurrent requests emit
verbatim-identical prefixes at identical decode positions. In one of our
own probe categories, 4 of 16 requests generated byte-identical 63-token
outputs. The resulting expert overlap is easily mistaken for semantic
locality: restricting the comparison to request pairs whose generated
text barely overlaps collapses one category's within-class expert
Jaccard from 0.229 to 0.092, against a cross-category baseline of 0.078.
We introduce a matched-pair design --- the same source records rendered
both with diverse and with fixed templates --- that changes the
early-window effect by 19.4--31.9 percentage points. An abrupt decay
aligned with a shared generated prefix is a diagnostic of surface
repetition, though genuine task effects may also vary by response phase.
Correcting the construction reverses the point-estimate ordering; only
two of six corrected category effects replicate clearly across two
request draws.

\emph{Operating regimes.} Normalized miss fractions are routinely
compared across MoE models with different expert counts. They should not
be. The relevant variable is the ratio of the per-step, per-layer expert
union to the per-layer cache capacity. As the fraction of events
exceeding capacity grows, capacity-forced traffic increases and policy
headroom can collapse --- so a small gap can mean ``the capacity floor
dominates'' rather than ``existing policies suffice''. Holding the model
and the workload fixed and changing only the cache fraction moves the
gap from 7.4\% to 33.3\%. The ratio is necessary but not sufficient:
permuting only the temporal order of an event stream, with the event set
and the ratio held exactly constant, moves the gap from 44.9\% to 30.8\%
and changes which causal policy is best.

\textbf{What remains after correction.} With all three controlled, a
large gap to the offline optimum persists: 44.2--45.9\% across 13 frozen
workload compositions at \ensuremath{\rho}=40\% and B=8, and 50.4\% at
B=2. The natural reading is that a substantial opportunity is waiting
for a better online policy. We tested that reading directly.

Belady's advantage over a causal policy has two sources: refusing to
admit a block that will not be reused before it would be evicted, and
knowing which resident block is used furthest in the future. Only the
first has a clear online analogue. Replacing the oracle with one that is
forced to admit every miss separates them: bypass admission accounts for
15.7\% of the gap at B=8 and 3.4\% at B=2, leaving 84.3\% and 96.6\% to
future-victim knowledge. We then trained a next-use-distance predictor
on causal features available at eviction time --- recency, frequency,
gate mass, concurrent routing multiplicity, layer, and popularity rate
--- fitted on one trace and evaluated on a disjoint one. It transfers at
R\ensuremath{^2}=0.24 and, substituted for the oracle's input in the
same eviction and admission machinery, recovers \textminus{}11.4\% of
the gap: it is worse than the causal baseline it was meant to improve.

\textbf{Position.} We state our claim narrowly. We do not show that
expert caching is a closed problem, that future-victim knowledge is
unpredictable in principle, or that online MoE cache controllers have no
future; existing work on expert \emph{prefetching} predicts a different
quantity --- which experts will be routed next, rather than how long a
cached block will remain unused --- and is not contradicted by our
result. What we show is that in our evaluated settings a large
offline-optimal gap substantially overstates the gains recovered by
representative lightweight causal mechanisms, and that reporting such a
gap without decomposing it invites an unwarranted inference. We
accompany this with three scoped negative results and one invalidated
mechanism experiment --- steady-state prefill eviction, warm-cache
workload transition, semantic cache partitioning, and causal affinity
batching --- each annotated with the model, cache fraction, concurrency,
scheduling discipline, and waiting constraint under which it was
measured, because we found that at least one of these mechanisms fails
for a reason that does not extrapolate.

\textbf{Contributions.}

\begin{enumerate}
\def\labelenumi{\arabic{enumi}.}
\tightlist
\item
  We formalize and implement an \textbf{event-atomic replay protocol}
  for the fused-event traffic contract used in our trace-driven
  evaluation, show that an inconsistent flattened replay selectively
  penalizes recency-based policies and inverts policy rankings, and
  provide a hand-verifiable reference trace (§2, §4).
\item
  We introduce a \textbf{matched-pair method} for quantifying
  prompt-template contamination in workload-conditioned routing studies,
  together with prompt-side and route-side diagnostics, and release a
  diversity-controlled probe set built to be free of it (§5).
\item
  We show that \textbf{cross-model comparison of MoE cache results
  requires explicitly controlling the operating regime}, give the
  union-to-capacity ratio as a first-order variable that must be
  reported, and demonstrate by ablation that it is necessary but not
  sufficient (§6).
\item
  We \textbf{decompose the offline-optimal gap} into an
  online-approximable and a future-dependent component and show, for
  representative causal mechanisms in our settings, a large gap between
  the oracle bound and realizable gains (§7, §8).
\item
  We distill the above into a \textbf{reporting checklist} for future
  MoE cache evaluations (§10) and will release, on publication, the
  simulator, probe set, contamination tools, frozen manifests and ---
  where upstream licences permit --- routing artifacts and
  pre-registration documents including failed criteria (§13).
\end{enumerate}

\textbf{Scope.} All results are trace-driven simulation; we do not
measure real host-to-device transfers, kernel time, or interconnect
topology. Our primary results use a single model, with two smaller
models for cross-model checks, and we evaluate only strictly lossless
policies --- no expert substitution, pruning, or reduced-precision
fallback. Section 11 states these limits in full. We regard the
checklist in §10, rather than any individual number, as the most
portable result of this work.

\textbf{Relation to prior evaluations.} We audited the public reporting
of ten representative papers (§12). Seven are end-to-end systems, so our
replay failure is not applicable to their actual execution; one is
trace-driven but does not report enough detail to reconstruct our
fused-event contract and assumes batch size one; two study architecture
or prediction targets. None directly reports the measured
union-to-capacity ratio needed to place its results in §6's regime, and
prompt-template multiplicity is generally not reported. These omissions
do not make the results incorrect. They prevent direct comparison under
a common evaluation contract, which is the motivation for §10's
checklist.

\section{§2 Evaluation Model}\label{evaluation-model}

Results in this area are difficult to compare because the underlying
replay model is rarely stated. We therefore fix notation and semantics
before reporting any number. Everything in §4--§8 is defined against
this model, and a hand-verifiable reference trace is released with the
artifacts.

\subsection{2.1 Traces and events}\label{traces-and-events}

A model has \texttt{L} MoE layers and \texttt{N} routed experts per
layer, and routes each token to \texttt{k} experts per layer. The unit
of cache management is an \textbf{expert block} \texttt{b\ =\ (l,\ e)}
with \texttt{l\ in\ {[}L{]}}, \texttt{e\ in\ {[}N{]}}; the block
universe has size \texttt{L.N}.

Requests are served under first-come-first-served continuous batching.
At scheduling step \texttt{s}, \texttt{A(s)} denotes the set of active
requests. For the fused-event traffic contract evaluated in this paper,
the quantity that matters for memory traffic is not the per-token expert
assignment but its \textbf{union over the concurrent batch}:

\begin{quote}
\textbf{Definition 1 (event).} An \emph{event} is a pair
\texttt{(s,\ l)}. Its expert set is
\texttt{E(s,\ l)\ =\ union\_\{r\ in\ A(s)\}\ topk(r,\ l,\ s)}, the
distinct expert blocks that step \texttt{s} touches at layer \texttt{l}.
\end{quote}

A \textbf{trace} is the sequence of events ordered lexicographically by
\texttt{(s,\ l)}. Deduplication inside an event is a property of the
execution unit, not of any cache policy: an expert selected by five
concurrent requests is fetched once. We therefore treat it as free and
never credit it to a policy (§2.4).

\subsection{2.2 Event-atomic replay}\label{event-atomic-replay}

\begin{quote}
\textbf{Definition 2 (event-atomic replay).} Let \texttt{C} be the cache
contents. For each event \texttt{(s,\ l)} in trace order:

\begin{enumerate}
\def\labelenumi{\arabic{enumi}.}
\tightlist
\item
  take the snapshot \texttt{C} at event start;
\item
  classify every member of \texttt{E(s,l)} against that snapshot:
  \texttt{hits\ =\ E\ intersect\ C},
  \texttt{misses\ =\ E\ \textbackslash{}\ C};
\item
  serve \textbf{all} of \texttt{misses} (no partial service);
\item
  only after the event completes, apply admission and eviction to obtain
  \texttt{C\textquotesingle{}}.
\end{enumerate}
\end{quote}

The ordering constraint is step 4. Admission and eviction are deferred
to the event boundary, so no member of \texttt{E\ intersect\ C} can be
evicted before the event that needs it has completed. This counts each
missing expert once and assumes the execution workspace can stream that
expert, compute all tokens routed to it, and release temporary state
before choosing the retained cache contents. It does \textbf{not} assume
that all members of \texttt{E} fit in the cache simultaneously.

\textbf{Sequential replay}, the alternative we examine in §4, flattens
\texttt{E(s,l)} into an ordered list of individual accesses and applies
admission and eviction after each one. When
\texttt{\textbar{}E(s,l)\textbar{}} exceeds the layer's quota, an access
late in the event can evict a block in \texttt{E\ intersect\ C} that the
\emph{same} event has not yet consumed, which is then counted as a miss.

\textbf{A remark on modelling.} We do not claim event-atomic replay is
the only admissible model of MoE execution. An engine that serializes
expert loads within a layer is a legitimate alternative, and a simulator
may reasonably model it. The distinction we draw is narrower and is a
correctness property, not a preference:

\begin{quote}
Sequential \emph{execution} is a realizable system. What is inconsistent
is to claim one-transfer-per-distinct-expert fused-event accounting
while evicting a block resident at event start, still required later in
that committed event, and counting its refetch as if it were a new
demand.
\end{quote}

A simulator should state which contract it targets. Section 4 measures
what happens when a flattened replay is used to estimate the fused-event
contract without protecting start-resident, not-yet-served members.

\subsection{2.3 Cache scope, capacity and
policies}\label{cache-scope-capacity-and-policies}

Capacity is a residency fraction \texttt{rho} of the block universe; the
resident set at any instant is the analogue of a working set in the
classical sense (\citeproc{ref-denning1968workingset}{Denning 1968}).
Under \textbf{per-layer scope} the budget is divided into quotas
\texttt{c\_l}, with the remainder to the lowest-indexed layers; under
\textbf{global scope} a single pool serves all layers. Per-layer scope
is primary throughout, because expert weights are consumed
layer-by-layer within a step.

We evaluate \texttt{Demand} (no cache), \texttt{Static} (a fixed set
chosen offline from a calibration trace and never updated),
\texttt{LRU}, \texttt{LFU}, \texttt{LFRU} (rank by
\texttt{frequency\ /\ age}), \texttt{Least-Stale} (blocks untouched in
the current forward cycle are evicted first), \texttt{Belady}
(\citeproc{ref-belady1966replacement}{Belady 1966}) (offline optimum
with bypass admission), and \texttt{Belady-forced-admit}, which is not
deployable but is the decomposition probe of §7.2. Ties break by seeded
random priority; §7.1 reports tie-seed sensitivity. Two static pin
protocols appear in this paper and are named where used: a
\textbf{calibration-fitted} list, frozen from a disjoint trace and used
in §9, and a \textbf{same-trace diagnostic} list used only in Table 2,
where the question is whether replay semantics moves a policy that by
construction cannot move. The initial load of a pinned set is counted in
transferred blocks and reported separately from steady-state traffic.

\subsection{2.4 Metrics}\label{metrics}

Let \texttt{M} be the number of blocks transferred over a trace, and
define

\begin{itemize}
\tightlist
\item
  \texttt{logical} --- expert assignments \textbf{before} intra-event
  deduplication. For the decode traces used in the primary experiments
  this is \texttt{sum\_\{s,l\}\ \textbar{}A(s)\textbar{}\ .\ k}; for
  prefill it must additionally sum the routed tokens contributed by each
  request rather than merely the request count;
\item
  \texttt{union\ \ \ =\ sum\_\{s,l\}\ \textbar{}E(s,l)\textbar{}} ---
  distinct blocks touched.
\end{itemize}

We report

\begin{quote}
\texttt{effective\ miss\ fraction\ \ m\ =\ M\ /\ logical}
\texttt{recoverable\ gap\ \ \ \ \ \ \ \ \ \ g\ =\ (m\_best-causal\ -\ m\_Belady)\ /\ m\_best-causal}
\end{quote}

and, separately, \texttt{union\ /\ logical}.

Two conventions deserve emphasis. First, \texttt{m} uses the
\textbf{pre-deduplication} denominator so that a single number remains
comparable across concurrency levels; the deduplication benefit itself
is reported as \texttt{union\ /\ logical} and is never counted as a
policy's contribution, because any FCFS scheduler obtains it for free.
Second, \texttt{g} is a \emph{relative} quantity and is meaningful only
alongside the absolute traffic; §6.1 shows that a small \texttt{g} can
indicate either that existing policies suffice or that no policy can
help, and that the two are distinguished only by absolute transferred
bytes against a bandwidth budget.

For a static pinned set, the initial load of the pinned blocks is
counted in \texttt{M} and is reported separately from steady-state
traffic. Because this one-time cost is sensitive to trace length,
static/dynamic comparisons must also report the fit source, quota rule,
number of events, and either total traffic or an amortized steady-state
view.

\subsection{2.5 Reference trace}\label{reference-trace}

The artifact includes a deliberately small trace --- 3 layers, 4 experts
per layer, 5 scheduling steps, 30 union accesses --- that can be
replayed by hand. Under Definition 2 with the released capacity setting
it yields 12 hits and 18 misses. It is intended as an executable
specification: a simulator that does not reproduce these counts does not
implement the semantics used in this paper.

\section{§3 Experimental Setup}\label{experimental-setup}

\subsection{3.1 Why expert caching is a cache problem with unusual
structure}\label{why-expert-caching-is-a-cache-problem-with-unusual-structure}

A routed MoE layer selects \texttt{k} of \texttt{N} experts per token;
under offloading the resident subset lives in device memory and the
remainder is streamed on touch. Three properties shape every measurement
that follows. The access unit is a set, not an element: all experts a
step touches at a layer are required together (§2.1), so hit/miss
classification is naturally event-level. Concurrency deduplicates for
free: experts selected by several concurrent requests are fetched once,
a benefit belonging to the scheduler rather than to any policy, which we
account for separately throughout. And the binding budget is bandwidth,
not capacity alone: raising the residency fraction lowers traffic but
consumes memory that also serves KV state and workspace, so a policy is
useful only if it reduces bytes at fixed residency, or permits lower
residency at fixed service quality.

\subsection{3.2 Models}\label{models}

{\def\LTcaptype{none} 
\begin{longtable}[]{@{}lrrrr@{}}
\toprule\noalign{}
& experts \texttt{N} & top-\texttt{k} & MoE layers \texttt{L} & blocks
\texttt{L.N} \\
\midrule\noalign{}
\endhead
\bottomrule\noalign{}
\endlastfoot
Granite-3.1-3B-A800M & 40 & 8 & 32 & 1,280 \\
OLMoE-1B-7B-0125 & 64 & 8 & 16 & 1,024 \\
\textbf{Qwen3-30B-A3B (4-bit)} & \textbf{128} & \textbf{8} & \textbf{48}
& \textbf{6,144} \\
\end{longtable}
}

The model-family descriptions follow their official technical reports or
model documentation (\citeproc{ref-ibm2024granite31}{IBM Research 2024};
\citeproc{ref-muennighoff2025olmoe}{{Muennighoff et al.} 2025};
\citeproc{ref-yang2025qwen3}{{Yang et al.} 2025}).

The exact model revisions are frozen in the route manifests: Granite
commit \texttt{d4dd87aa3a6c201bc374851d7d7ff4cf39a0b82a}, OLMoE commit
\texttt{9b0c1aa87e34a20052389dce1f0cf01da783f654}, and the MLX Qwen3
snapshot \texttt{d388dead1515f5e085ef7a0431dd8fadf0886c57} (4-bit, group
size 64).

Qwen3-30B-A3B is the primary model: its expert count places it at the
boundary where the per-step expert union crosses per-layer cache
capacity (§6), which is the regime the two smaller models cannot reach.
The smaller models are used for cross-model checks and for the
replay-semantics measurement of §4. All three are evaluated at the same
residency fractions and under identical replay semantics.

\textbf{Analytical reference frame.} To express results in engineering
units we also use a paper specification of a frontier-scale MoE: 92 MoE
layers, 896 routed experts, top-16, 82,432 expert blocks of
approximately 16.74 MiB, approximately 106.55 GiB of non-routed weights,
approximately 1,347.12 GiB of routed expert weights, and approximately
25.83 GB of logical expert bytes per output token before deduplication.
\textbf{No trace from this model was collected and no measurement in
this paper was taken on it.} It is used only to state operating regimes
and unit conversions, and every quantity derived from it is labelled as
an analytical projection, never as a prediction.

\subsection{3.3 Traces}\label{traces}

Decode routing was collected autoregressively with the model's chat
template applied and reasoning mode disabled, greedy decoding, prompts
capped at 4,096 tokens, and up to 384 recorded decode forwards per
request with natural end-of-sequence termination. For each request we
store the emitted token sequence, the effective length, and the complete
per-layer router logits, top-\texttt{k} indices, probabilities, gate
weights, entropy and margin.

Two collection choices matter for later sections and are the direct
consequence of §5. Applying the chat template prevents the model from
continuing the instruction text instead of answering it. Recording
several hundred decode steps places the majority of each trace beyond
the boilerplate-dominated opening region. An earlier collection that did
neither is retained and re-analysed in §5 as the contaminated condition.

Traces are converted into event streams (§2.1) under
first-come-first-served continuous batching with same-step cross-request
deduplication. Admission offsets are configurable so that concurrent
requests are not locked to identical decode positions.

The primary Qwen3 collection ran on a Mac mini with an Apple M4 Pro and
24 GiB unified memory under macOS 26.5.1, Python 3.12.13, MLX 0.32.0 and
MLX-LM 0.31.3. It contains 144 confirmatory requests in 24 shards, uses
the snapshot above, and reached 19.31 GB peak Metal allocation. All
collection parameters and shard hashes are stored in the source manifest
rather than inferred from filenames. MLX is cited as versioned software
rather than as an inferred hardware-system paper
(\citeproc{ref-mlx2025}{MLX Contributors 2025}).

\subsection{3.4 Workload probe set}\label{workload-probe-set}

Measuring workload-conditioned routing requires requests partitioned by
task. We use \textbf{ControllerProbe-D1}: 432 records over six workload
archetypes --- document RAG, tool agent, ERP structured analytics,
office/legal, process diagnostics, and equipment maintenance/BOM ---
built from public benchmark and public industrial sources. The
benchmark-derived portions use LongBench, Berkeley Function Calling
Leaderboard, Spider 2.0, and LegalBench
(\citeproc{ref-bai2023longbench}{Bai et al. 2023};
\citeproc{ref-yan2024bfcl}{{Yan et al.} 2024};
\citeproc{ref-lei2024spider20}{{Lei et al.} 2024};
\citeproc{ref-guha2023legalbench}{{Guha et al.} 2023}); the industrial
portions include the Petrobras 3W data and Apache OFBiz sample entities
(\citeproc{ref-vargas2019threew}{Vargas et al. 2019};
\citeproc{ref-apacheofbiz2026}{Apache Software Foundation 2026}), with
the complete source and licence ledger released alongside the artifact.
Its construction is described in §5; the properties that matter here
are:

\begin{itemize}
\tightlist
\item
  twelve structurally distinct prompt forms, each terminating in a
  different kind of line, crossed with two instruction languages, so
  that no two records in a cell share a form--language pair;
\item
  eight task framings per archetype and five payload rendering styles;
\item
  a \textbf{matched-pair control arm} in which a subset of the same
  source records is re-rendered with a single fixed template, so that
  template effects can be measured rather than argued;
\item
  a group-aware discovery/confirmatory split with no source group in
  both sides.
\end{itemize}

Category proportions in the probe set are a design choice for controlled
comparison and are \textbf{not} an estimate of any deployment's traffic
mix.

\subsection{3.5 Simulator and statistical
discipline}\label{simulator-and-statistical-discipline}

The simulator implements Definition 2 with per-layer and global scope,
seeded tie-breaking, and the policy set of §2.3. It is validated against
the reference trace of §2.5.

Every threshold presented as a confirmatory \textbf{decision criterion}
was pre-registered and frozen before the corresponding measurement was
read, together with the split it would be evaluated on. Exploratory
diagnostics, failure attribution, and reviewer-driven audits were not
pre-registered. We release the decision documents unchanged, including
those in which a criterion failed and the associated development line
was stopped. Where a design parameter was revised, the revision and its
rationale are recorded with the statement that no result had yet been
generated or read; §8 gives one such case. Discovery and confirmatory
splits are disjoint at the source-group level, and confirmatory data is
read once.

We report tie-seed sensitivity for the principal conditions. We do not
report population confidence intervals for quantities estimated from a
single trace, and we mark such quantities as implementation-stable
rather than statistically bounded.

All simulation, workload construction and analysis code released with
this paper was implemented with the assistance of Codex (OpenAI) and
Claude (Anthropic) under the author's direction, and the intermediate
analyses from which the author worked were produced with the same
assistance. Correctness was established by regenerating every reported
figure from the released code against frozen result manifests with
recorded SHA-256 hashes, rather than by inspection of the generated code
alone. The verification procedure and the manifests are part of the
released artifacts, so any reader may repeat the check independently.
The declaration at the end of this paper states the assistance in full.

\section{§4 Axis I --- Replay Semantics}\label{axis-i-replay-semantics}

\begin{quote}
\textbf{Claim.} Under the fused-event traffic contract of §2, replaying
an event as individual accesses permits eviction and artificial refetch
of an expert that was resident at event start and is still required
later in the same event. The resulting error is not uniform noise: it
penalizes recency-based policies by roughly 27--29\% while leaving
frequency-based, offline, and static policies within 4\%, and it inverts
the policy ranking.
\end{quote}

\subsection{4.1 The mechanism}\label{the-mechanism}

Under sequential replay, early misses may be admitted before all
start-resident members of the event have been served. If an early
admission evicts a resident expert that appears later in the flattened
order, that later access is counted as a miss even though the
fused-event contract would have counted it as a hit. Evicting an
already-served expert does not create a current-event miss; it can,
however, bias the event's final retained set and therefore future
events.

A recency-based policy is especially sensitive because the arbitrary
order inside the event immediately changes its key and its victims.
Frequency-based policies are less sensitive when their counters are
dominated by prior events. The offline optimum uses future position, and
a static pinned set never evicts. This difference in policy sensitivity,
rather than a uniform counting offset, is the mechanism tested below.

Expert identifiers make this worse in a specific way. If the flattened
order is the natural one --- ascending expert index --- then the
surviving residents are systematically the highest-indexed experts of
each event, and expert numbering, an arbitrary artifact of the
checkpoint, enters the retention decision.

\subsection{4.2 Randomization is not a
fix}\label{randomization-is-not-a-fix}

A natural repair is to randomize the intra-event order. Table 1 shows
that this addresses only half of the problem.

\textbf{Table 1.} Miss ratio under three replay semantics.
\emph{(64-expert model, per-layer scope, \ensuremath{\rho} = 40\%, LRU;
range over 5 tie seeds.)}

{\def\LTcaptype{none} 
\begin{longtable}[]{@{}lr@{}}
\toprule\noalign{}
replay semantics & miss ratio \\
\midrule\noalign{}
\endhead
\bottomrule\noalign{}
\endlastfoot
sequential, ascending expert index & 99.73\% \\
sequential, seeded-random order & 87.15\% -- 87.19\% \\
\textbf{event-atomic (Definition 2)} & \textbf{56.92\% -- 57.11\%} \\
\end{longtable}
}

Randomization removes the dependence on expert numbering, which accounts
for the drop from 99.73\% to roughly 87\%. It does not remove
intra-event eviction of not-yet-consumed residents, which accounts for
the remaining 30 percentage points. Only deferring the retention
decision to the event boundary removes both.

\subsection{4.3 The distortion is selective, and it inverts
rankings}\label{the-distortion-is-selective-and-it-inverts-rankings}

Table 2 holds the trace, capacity, concurrency, cache scope and
residency fraction fixed and varies only the replay semantics. It is the
data behind Figure 1.

\textbf{Table 2.} Effective miss fraction under sequential and
event-atomic replay. \emph{(128-expert model, held-out split, per-layer
scope, \ensuremath{\rho} = 40\%, B = 8. Identical trace and capacity in
both columns.)}

{\def\LTcaptype{none} 
\begin{longtable}[]{@{}lrrr@{}}
\toprule\noalign{}
policy & sequential replay & event-atomic & inflation \\
\midrule\noalign{}
\endhead
\bottomrule\noalign{}
\endlastfoot
LRU & 24.59\% & 19.30\% & +27.4\% \\
LFRU & 23.14\% & 18.01\% & +28.5\% \\
Least-Stale & 24.59\% & 19.30\% & +27.4\% \\
LFU & 19.80\% & 19.14\% & +3.4\% \\
Belady & 10.36\% & 9.93\% & +4.3\% \\
Static-same-trace (diagnostic) & 19.10\% & 19.10\% & 0.0\% \\
\end{longtable}
}

\begin{figure}
\centering
\includegraphics[width=0.95\linewidth,height=\textheight,keepaspectratio,alt={Figure 1. Replay semantics selectively changes policy measurements. The trace, capacity, concurrency, scope and residency fraction are identical; only replay semantics changes. Values are built from the frozen Table 2 artifact.}]{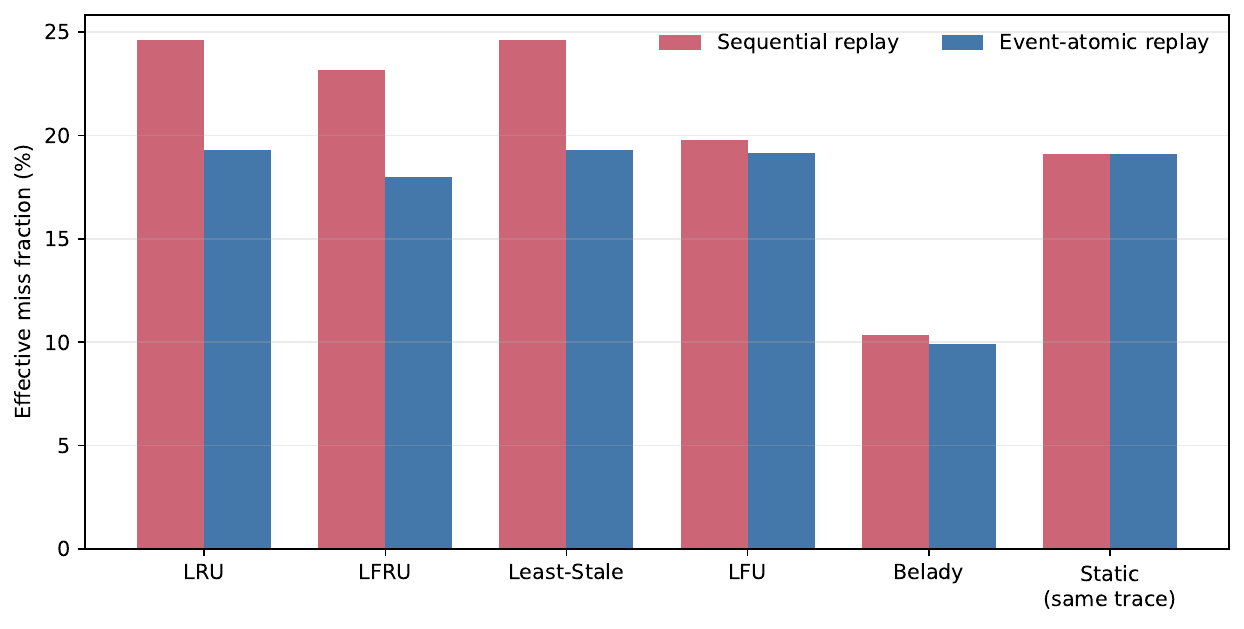}
\caption{\textbf{Figure 1. Replay semantics selectively changes policy
measurements.} The trace, capacity, concurrency, scope and residency
fraction are identical; only replay semantics changes. Values are built
from the frozen Table 2 artifact.}\label{fig:replay-semantics}
\end{figure}

Two observations follow.

\textbf{The ranking inverts.} Under sequential replay the best causal
policy is the same-trace static diagnostic, which appears to beat LFRU
by 4.04 percentage points. Under event-atomic replay LFRU is best and
beats that diagnostic by 5.70\% relative (1.09 percentage points). A
study that concluded ``a fixed, offline-chosen residency set outperforms
online replacement'' would be an artifact of the replay model alone.

The static row is fitted on the evaluated trace in both columns so that
replay semantics is the only changed variable. It is therefore evidence
of replay-invariance, not evidence for a deployable static policy; §9
evaluates the discovery-frozen protocol separately.

\textbf{The error cannot be treated as a calibration offset.} Because
the inflation ranges from 0.0\% to 28.5\% across policies, no single
correction factor recovers the correct ordering. Results obtained under
sequential replay are not convertible to event-atomic results after the
fact.

Least-Stale coincides exactly with LRU in both columns. This is expected
rather than anomalous: in the absence of prefetching, the two-generation
staleness test degenerates to last-access order. We retain it as a
distinct implementation so that the diagnostic is explicit, and we note
that its published advantage is reported in combination with prefetching
and miss handling, which we do not evaluate here.

\subsection{4.4 Consequence for a downstream
conclusion}\label{consequence-for-a-downstream-conclusion}

The semantics changed one of our own conclusions rather than merely its
precision. Under event-atomic replay, the steady-state prefill gap
between the best causal policy and the offline optimum is:

\textbf{Table 3.} Recoverable gap, prefill, per-layer scope, mean over 5
tie seeds.

{\def\LTcaptype{none} 
\begin{longtable}[]{@{}
  >{\raggedright\arraybackslash}p{(\linewidth - 6\tabcolsep) * \real{0.2000}}
  >{\raggedleft\arraybackslash}p{(\linewidth - 6\tabcolsep) * \real{0.2667}}
  >{\raggedleft\arraybackslash}p{(\linewidth - 6\tabcolsep) * \real{0.2667}}
  >{\raggedleft\arraybackslash}p{(\linewidth - 6\tabcolsep) * \real{0.2667}}@{}}
\toprule\noalign{}
\begin{minipage}[b]{\linewidth}\raggedright
model
\end{minipage} & \begin{minipage}[b]{\linewidth}\raggedleft
\ensuremath{\rho} = 20\%
\end{minipage} & \begin{minipage}[b]{\linewidth}\raggedleft
\ensuremath{\rho} = 30\%
\end{minipage} & \begin{minipage}[b]{\linewidth}\raggedleft
\ensuremath{\rho} = 40\%
\end{minipage} \\
\midrule\noalign{}
\endhead
\bottomrule\noalign{}
\endlastfoot
40-expert & 0.155\% & 0.320\% & 0.733\% \\
64-expert & 0.601\% & 1.516\% & 3.169\% \\
\end{longtable}
}

All six points fall far below the 10\% threshold we had pre-registered
as the level at which designing a new steady-state prefill eviction rule
would be worthwhile. Under the earlier sequential semantics the same
measurements had appeared to leave several times that much headroom, and
a conclusion we had drawn from them --- that admission and pinning were
the dominant mechanism --- did not survive the correction and was
formally withdrawn.

We report this because it illustrates the failure mode the paper is
about. The sequential result was not obviously wrong when produced: the
miss ratios were plausible, the policies were implemented correctly, and
the relative ordering looked reasonable. The defect was in the replay
model, which is exactly the component least often described.

The magnitude of the inflation is not a constant: it grows with
concurrency and shrinks with residency fraction, converging when events
fit inside the quota. The qualitative property we claim --- selectivity
by policy family, and therefore ranking change --- follows from the
contract argument of §2.2 and does not depend on the operating point
(§11).

\section{§5 Axis II --- Workload
Contamination}\label{axis-ii-workload-contamination}

\begin{quote}
\textbf{Claim.} Probe sets that wrap each workload category in a single
instruction template produce verbatim-identical generation prefixes
across concurrent requests. The resulting expert overlap is readily
mistaken for semantic locality. A matched-pair rendering intervention
changes the measured early-window effect by 19.4--31.9 percentage
points, and correcting the construction reverses the point-estimate
category ordering.
\end{quote}

\subsection{5.1 A conclusion that did not
survive}\label{a-conclusion-that-did-not-survive}

We first measured workload-conditioned expert locality on a probe set
built the obvious way: six workload archetypes, one instruction template
per archetype, public payloads substituted into each. The result was
clean, passed a pre-registered confirmatory split, and had an appealing
interpretation --- industrial process-diagnostics traffic appeared to
concentrate on a markedly narrower expert working set than general
office traffic.

Table 4 shows what happened when the probe set was rebuilt to control
for prompt-surface repetition, holding the model, the payload sources,
the cache configuration, and the metric fixed.

\textbf{Table 4.} Reduction in effective miss fraction of a homogeneous
single-archetype stream relative to a mixed stream. \emph{(128-expert
model, per-layer scope, \ensuremath{\rho} = 40\%, B = 8, event-atomic.)}

{\def\LTcaptype{none} 
\begin{longtable}[]{@{}
  >{\raggedright\arraybackslash}p{(\linewidth - 4\tabcolsep) * \real{0.2727}}
  >{\raggedleft\arraybackslash}p{(\linewidth - 4\tabcolsep) * \real{0.3636}}
  >{\raggedleft\arraybackslash}p{(\linewidth - 4\tabcolsep) * \real{0.3636}}@{}}
\toprule\noalign{}
\begin{minipage}[b]{\linewidth}\raggedright
workload archetype
\end{minipage} & \begin{minipage}[b]{\linewidth}\raggedleft
single-template probe set
\end{minipage} & \begin{minipage}[b]{\linewidth}\raggedleft
diversity-controlled probe set
\end{minipage} \\
\midrule\noalign{}
\endhead
\bottomrule\noalign{}
\endlastfoot
process diagnostics & \textbf{57.2\%} \emph{(headline result)} &
\textbf{5.9\%} \\
document RAG & 57.9\% & 43.7\% \\
ERP structured analytics & 36.2\% & 7.6\% \\
equipment maintenance / BOM & 35.3\% & 22.0\% \\
office / legal & 26.9\% & \textbf{53.3\%} \\
tool agent & 9.5\% & 10.2\% \\
\end{longtable}
}

The point-estimate ordering does not merely compress; it inverts. The
archetype that had supported the strongest claim falls to near the
bottom, and the archetype that had ranked second from last rises to the
top. We withdrew the original conclusion. The remainder of this section
explains why it was wrong, how to detect the failure, and how much of
the corrected ordering is actually supported.

\subsection{5.2 Mechanism}\label{mechanism}

Three conditions combine, none of which is individually unusual.

\emph{A shared template tail.} Every prompt in a category ends with the
same instruction sentence. Under raw continuation --- that is, when the
model's chat template is not applied --- the model completes that
sentence verbatim before producing anything task-specific, and then
emits a fixed reasoning preamble.

\emph{A short decode window.} If only a few dozen decode steps are
recorded, the shared prefix is a large fraction of every trace. In our
original collection the first 15--25 of 63 recorded steps were
near-identical across requests.

\emph{Position-locked cohorts.} If all requests are the same length and
are admitted and retired in lock-step, concurrent requests occupy
identical decode positions, so the shared prefix aligns exactly at the
same scheduling step. The per-step expert union is then computed over
near-identical token contexts.

The consequence is direct. In one archetype, 4 of 16 requests produced
byte-identical 63-token outputs; positional token agreement within that
archetype was 20.65\%, against 1.04\% for a category whose payloads were
long heterogeneous documents.

\subsection{5.3 Two diagnostics that separate the
explanations}\label{two-diagnostics-that-separate-the-explanations}

Within-category expert overlap alone cannot distinguish semantic
locality from surface repetition. Two measurements do, and neither
requires a control arm.

\emph{Residual after removing near-duplicate text.} We restrict the
comparison to request pairs whose generated text barely overlaps ---
positional token agreement below 5\% --- and compare the residual
against a cross-archetype baseline.

\textbf{Table 5.} Within-archetype pairwise expert Jaccard, all pairs
and restricted to low-text-overlap pairs. \emph{(16 requests per
archetype.)}

{\def\LTcaptype{none} 
\begin{longtable}[]{@{}
  >{\raggedright\arraybackslash}p{(\linewidth - 8\tabcolsep) * \real{0.1579}}
  >{\raggedleft\arraybackslash}p{(\linewidth - 8\tabcolsep) * \real{0.2105}}
  >{\raggedleft\arraybackslash}p{(\linewidth - 8\tabcolsep) * \real{0.2105}}
  >{\raggedleft\arraybackslash}p{(\linewidth - 8\tabcolsep) * \real{0.2105}}
  >{\raggedleft\arraybackslash}p{(\linewidth - 8\tabcolsep) * \real{0.2105}}@{}}
\toprule\noalign{}
\begin{minipage}[b]{\linewidth}\raggedright
\end{minipage} & \begin{minipage}[b]{\linewidth}\raggedleft
single-template set
\end{minipage} & \begin{minipage}[b]{\linewidth}\raggedleft
\end{minipage} & \begin{minipage}[b]{\linewidth}\raggedleft
diversity-controlled set
\end{minipage} & \begin{minipage}[b]{\linewidth}\raggedleft
\end{minipage} \\
\midrule\noalign{}
\endhead
\bottomrule\noalign{}
\endlastfoot
archetype & token agr. & expJac \textbar{} lowTok & token agr. & expJac
\textbar{} lowTok \\
process diagnostics & 20.65\% & 0.158 & 0.78\% & 0.087 \\
equipment / BOM & 15.33\% & \textbf{0.092} & 1.06\% & 0.108 \\
ERP & 5.15\% & 0.118 & 0.61\% & 0.096 \\
tool agent & 1.85\% & 0.103 & 0.53\% & 0.085 \\
document RAG & 1.04\% & 0.144 & 0.45\% & 0.119 \\
office / legal & 0.74\% & 0.123 & 0.57\% & 0.144 \\
\emph{cross-archetype baseline} & & \emph{0.078} & & \emph{0.077} \\
\end{longtable}
}

On the single-template set, equipment/BOM has a raw within-class Jaccard
of 0.229; restricting to low-text-overlap pairs collapses it to 0.092,
essentially the cross-archetype baseline of 0.078, and process
diagnostics falls from 0.291 to 0.158. Categories whose payloads were
naturally heterogeneous barely move. On the diversity-controlled set all
sequences are unique and the two columns nearly coincide, indicating
that the surface-repetition channel has been closed rather than reduced.
We report \texttt{expJac\ \textbar{}\ lowTok} minus the cross-archetype
baseline as the residual workload effect.

\emph{Prefix-aligned decay.} Template echo must fade as the shared
prefix ends. A genuine task effect can also vary across response phases,
so position dependence alone is neither necessary nor sufficient; the
stronger signal is an abrupt decay aligned with a shared, near-identical
generated prefix.

\textbf{Table 6.} Reduction in per-step expert union of a homogeneous
stream relative to a category-round-robin mixture, by decode-position
band. \emph{(B = 16.)}

{\def\LTcaptype{none} 
\begin{longtable}[]{@{}lrrrr@{}}
\toprule\noalign{}
archetype & single-template set & & diversity-controlled set & \\
\midrule\noalign{}
\endhead
\bottomrule\noalign{}
\endlastfoot
& early (t 0--8) & late (t 32--63) & early (t 0--16) & late (t
160--384) \\
process diagnostics & 65.7\% & 29.1\% & 6.0\% &
\textbf{\textminus{}7.8\%} \\
equipment / BOM & 39.2\% & 13.8\% & 3.7\% & 6.2\% \\
ERP & 24.5\% & 11.5\% & 6.8\% & \textminus{}8.6\% \\
document RAG & 1.2\% & \textbf{20.2\%} & 7.1\% & \textbf{17.1\%} \\
office / legal & 4.0\% & 9.7\% & 11.6\% & \textbf{20.1\%} \\
tool agent & 7.1\% & 5.8\% & \textminus{}8.6\% & 24.3\% \\
\end{longtable}
}

On the contaminated set the three template-rendered categories decay
steeply while the two with heterogeneous natural-language payloads are
flat or rising; on the corrected set the decaying categories collapse to
near zero or below and the rising ones persist. Combined with prompt and
output overlap this gives a screening rule: \textbf{an effect
concentrated in a position band occupied by a shared generated prefix
should be treated as surface repetition until a diverse or matched
control says otherwise.}

\subsection{5.5 Matched-pair
attribution}\label{matched-pair-attribution}

Diagnostics indicate contamination; they do not quantify it. We
therefore render a subset of the \emph{same source records} twice ---
once with the diversified forms and once with a single fixed template
ending in a fixed instruction --- and pair them by source record. Model,
cache configuration, concurrency and metric are identical; the
prompt-rendering policy is the treatment.

\textbf{Table 7.} Difference in apparent union reduction under the fixed
rendering, matched on source record, early decode band (t 0--16).

{\def\LTcaptype{none} 
\begin{longtable}[]{@{}lr@{}}
\toprule\noalign{}
archetype & template-attributable \\
\midrule\noalign{}
\endhead
\bottomrule\noalign{}
\endlastfoot
office / legal & +31.9 pp \\
process diagnostics & +28.8 pp \\
ERP & +26.4 pp \\
tool agent & +24.5 pp \\
equipment / BOM & +20.9 pp \\
document RAG & +19.4 pp \\
\end{longtable}
}

\begin{figure}
\centering
\includegraphics[width=0.92\linewidth,height=\textheight,keepaspectratio,alt={Figure 2. Prompt rendering contributes materially to the apparent early-window workload effect. Each bar is the fixed-template arm minus the diverse arm for the same source records at t=0--16; it is not an end-to-end miss reduction.}]{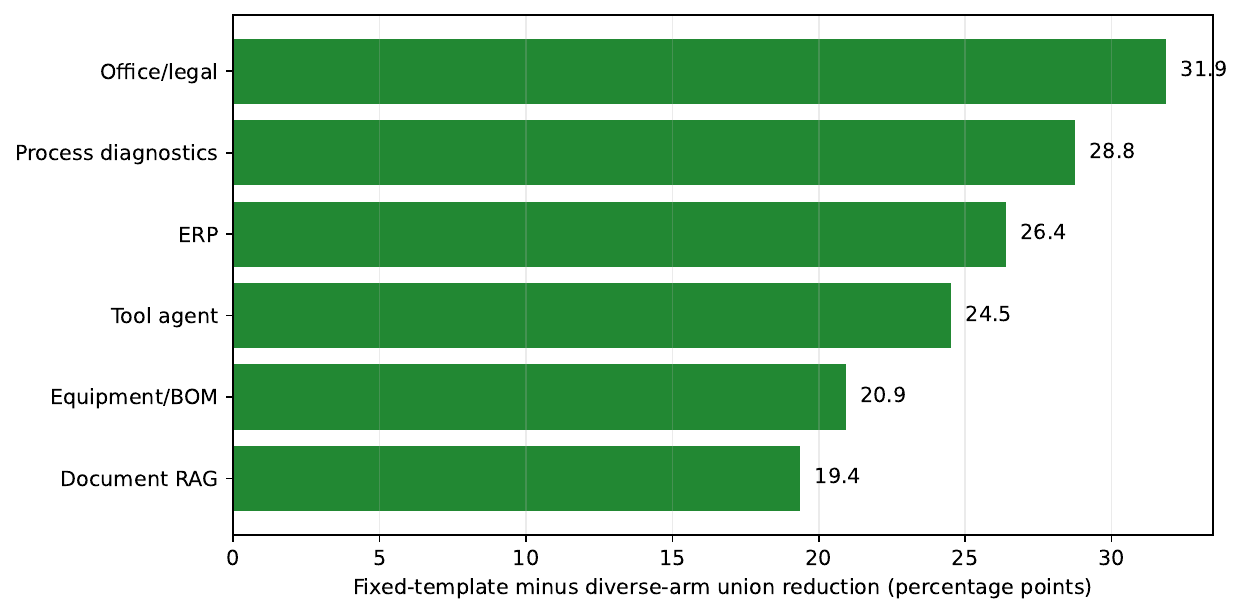}
\caption{\textbf{Figure 2. Prompt rendering contributes materially to
the apparent early-window workload effect.} Each bar is the
fixed-template arm minus the diverse arm for the same source records at
t=0--16; it is not an end-to-end miss
reduction.}\label{fig:template-attribution}
\end{figure}

This matched contrast is causal for the \textbf{rendering policy}, not
for a pure full-service template-echo component: changing the wrapper
also changes the continuation, response length and cache churn. Its
early position-aligned band, together with the shared prefix, is what
specifically supports the echo interpretation.

One asymmetry deserves note. The fixed-template arm has a \emph{higher}
effective miss fraction over the full trace (21.99\% vs 18.78\% at
\ensuremath{\rho} = 40\%, B = 8) even though it shows a larger
early-window union reduction, because the fixed template also shortens
responses and increases cold-start frequency. Template echo therefore
cannot be read off the end-to-end miss difference; only the
position-resolved union comparison isolates it.

\subsection{5.6 A diversity-controlled probe
set}\label{a-diversity-controlled-probe-set}

The corrected set holds payload sources fixed and varies only the
wrapper: twelve structurally distinct prompt forms, each terminating in
a different kind of line, crossed with two instruction languages, so
that no two records in an (archetype, split) cell share a form-language
pair; eight task framings per archetype; five payload rendering styles;
and, for the industrial archetypes, six source families with different
record shapes. Candidate payloads whose 5-gram Jaccard against an
already-selected payload exceeds 0.30 are rejected at selection time,
because several public sources are internally repetitive. The
matched-pair arm of §5.5 is built in. The result is 23-24 distinct
70-character prompt heads per 24-record cell against 1-4 per 20-record
cell in the original, and a mean pairwise 5-gram Jaccard of 0.002-0.014
against 0.038-0.190.

\subsection{5.7 How much of the corrected ordering is
supported}\label{how-much-of-the-corrected-ordering-is-supported}

Correcting a measurement invites the assumption that the corrected
result is sound. We checked. Two issues affect Table 4's right-hand
column.

First, the mixed reference stream contains six times as many distinct
requests as each homogeneous stream, and a stream with fewer distinct
requests has more inter-request reuse for reasons unrelated to workload.
We rebuilt the reference as seven size-matched draws. The confound
proves small --- the size-matched reference is 17.81\% against 18.76\%
for the original --- but the seven draws span 15.54\%--19.04\%, a
standard deviation of 1.18 percentage points.

Second, each homogeneous condition is itself a single draw of twelve
requests. We replicated all six on a disjoint held-out set of twelve.

\textbf{Table 8.} Homogeneous-stream miss fraction, two independent
draws, against a size-matched reference. Interval propagates the
reference's draw-to-draw spread; it is not a population confidence
interval.

{\def\LTcaptype{none} 
\begin{longtable}[]{@{}
  >{\raggedright\arraybackslash}p{(\linewidth - 10\tabcolsep) * \real{0.1364}}
  >{\raggedleft\arraybackslash}p{(\linewidth - 10\tabcolsep) * \real{0.1818}}
  >{\raggedleft\arraybackslash}p{(\linewidth - 10\tabcolsep) * \real{0.1818}}
  >{\raggedleft\arraybackslash}p{(\linewidth - 10\tabcolsep) * \real{0.1818}}
  >{\raggedleft\arraybackslash}p{(\linewidth - 10\tabcolsep) * \real{0.1818}}
  >{\raggedright\arraybackslash}p{(\linewidth - 10\tabcolsep) * \real{0.1364}}@{}}
\toprule\noalign{}
\begin{minipage}[b]{\linewidth}\raggedright
archetype
\end{minipage} & \begin{minipage}[b]{\linewidth}\raggedleft
draw 1
\end{minipage} & \begin{minipage}[b]{\linewidth}\raggedleft
draw 2
\end{minipage} & \begin{minipage}[b]{\linewidth}\raggedleft
mean
\end{minipage} & \begin{minipage}[b]{\linewidth}\raggedleft
improvement vs 17.81\%
\end{minipage} & \begin{minipage}[b]{\linewidth}\raggedright
\ensuremath{\pm}1 s.d. of reference
\end{minipage} \\
\midrule\noalign{}
\endhead
\bottomrule\noalign{}
\endlastfoot
office / legal & 8.76\% & 8.15\% & 8.45\% & \textbf{52.5\%} & 49.2 --
55.5 \\
document RAG & 10.57\% & 10.69\% & 10.63\% & \textbf{40.3\%} & 36.1 --
44.0 \\
equipment / BOM & 14.64\% & 13.33\% & 13.99\% & 21.5\% & 15.9 -- 26.4 \\
ERP & 17.33\% & 15.18\% & 16.25\% & 8.8\% & 2.3 -- 14.4 \\
process diagnostics & 17.65\% & 16.34\% & 16.99\% & 4.6\% &
\textbf{\textminus{}2.2 -- 10.5} \\
tool agent & 16.85\% & 18.17\% & 17.51\% & 1.7\% &
\textbf{\textminus{}5.3 -- 7.8} \\
\end{longtable}
}

Only the top two archetypes are clearly separated from the reference;
equipment/BOM is positive with a wide interval; and the bottom three are
not distinguishable from zero at this sample size. The honest statement
is therefore not that six archetypes form an ordering, but that
\textbf{two of six show a workload effect that survives both the
contamination correction and a size-matched reference.}

\subsection{5.8 Applicability beyond our own
data}\label{applicability-beyond-our-own-data}

The diagnostics of §5.3--§5.4 are not specific to our probe set. Applied
to the first 48 complete records of a public Mixtral routing artifact
over C4 input sequences, each truncated to 512 input tokens, they report
48 of 48 unique token sequences and a 95th-percentile positional
agreement of 1.37\% --- no contamination signal
(\citeproc{ref-raffel2020t5}{Raffel et al. 2020};
\citeproc{ref-allenai2024mixtralroutes}{Allen Institute for AI 2024}).
As a labelled sensitivity check, we then copied one record's first 64
token-and-route positions into every record. Positional agreement rose
from 0.641\% to 13.050\%, mean expert Jaccard from 0.185 to 0.286, and
the diagnostic flag changed from false to true.

The natural-C4 run is a false-positive check and the injected run is a
synthetic positive control. Neither is an external replication of the
prompt-rendering effect: the public artifact contains no workload labels
or matched template arm, and the injection is not a model-generated
response.

The effect requires a shared template tail, a decode window short enough
for the prefix to dominate, and, for the union metric, position-aligned
concurrency; we do not claim every template-based probe set satisfies
them (§11).

\section{§6 Axis III --- Operating
Regimes}\label{axis-iii-operating-regimes}

\begin{quote}
\textbf{Claim.} Normalized miss fractions do not transfer across MoE
models. The distribution of the per-step, per-layer expert union
relative to cache capacity is a first-order variable that must be
reported before any cross-model comparison is meaningful. Its mean is
necessary but not sufficient: holding the event-set multiset and mean
ratio exactly constant and permuting only temporal order moves the
offline-optimal gap from 44.9\% to 30.8\% and changes which causal
policy is best.
\end{quote}

\subsection{6.1 The regime variable}\label{the-regime-variable}

Two quantities determine whether a replacement policy has room to act at
all: how many distinct expert blocks a single event requires, and how
many the layer can hold. Write

\begin{quote}
\texttt{r(s,l)\ =\ \textbar{}E(s,l)\textbar{}\ /\ c\_l}, and
\texttt{r\_bar} is its mean over reported events.
\end{quote}

For an event with \texttt{r(s,l)\ \textless{}\ 1}, its working set fits
in the quota; misses can still arise from drift between events and
eviction order may matter. For an event with
\texttt{r(s,l)\ \textgreater{}\ 1}, at least
\texttt{\textbar{}E\textbar{}-c\_l} members cannot remain resident after
the event, creating a capacity-forced floor. As the mass of the
\texttt{r(s,l)} distribution above one grows, replacement policies have
less headroom. This is a tendency, not a theorem that all policies or
all traces converge.

\texttt{r\_bar} is not a free parameter. It generally rises with
concurrency and top-\texttt{k}, and falls with expert count and
residency fraction, so two models compared ``at the same batch size''
are routinely in different regimes.

\subsection{6.2 Crossing the boundary}\label{crossing-the-boundary}

Table 9 sweeps concurrency on the 128-expert model at fixed residency,
reporting the measured union, the fraction of events that exceed the
layer quota, and the resulting gap.

\textbf{Table 9.} Regime and gap versus concurrency. \emph{(128-expert
model, per-layer scope, \ensuremath{\rho} = 40\%, \texttt{c\_l}
approximately 51.2, event-atomic.)}

{\def\LTcaptype{none} 
\begin{longtable}[]{@{}
  >{\raggedleft\arraybackslash}p{(\linewidth - 10\tabcolsep) * \real{0.1667}}
  >{\raggedleft\arraybackslash}p{(\linewidth - 10\tabcolsep) * \real{0.1667}}
  >{\raggedleft\arraybackslash}p{(\linewidth - 10\tabcolsep) * \real{0.1667}}
  >{\raggedleft\arraybackslash}p{(\linewidth - 10\tabcolsep) * \real{0.1667}}
  >{\raggedleft\arraybackslash}p{(\linewidth - 10\tabcolsep) * \real{0.1667}}
  >{\raggedleft\arraybackslash}p{(\linewidth - 10\tabcolsep) * \real{0.1667}}@{}}
\toprule\noalign{}
\begin{minipage}[b]{\linewidth}\raggedleft
B
\end{minipage} & \begin{minipage}[b]{\linewidth}\raggedleft
union / layer-step
\end{minipage} & \begin{minipage}[b]{\linewidth}\raggedleft
p95
\end{minipage} & \begin{minipage}[b]{\linewidth}\raggedleft
\texttt{r\_bar}
\end{minipage} & \begin{minipage}[b]{\linewidth}\raggedleft
events over quota
\end{minipage} & \begin{minipage}[b]{\linewidth}\raggedleft
recoverable gap
\end{minipage} \\
\midrule\noalign{}
\endhead
\bottomrule\noalign{}
\endlastfoot
1 & 8.00 & 8 & 0.156 & 0.00\% & 46.97\% \\
2 & 15.02 & 16 & 0.293 & 0.00\% & 49.25\% \\
4 & 26.58 & 30 & 0.519 & 0.00\% & 48.12\% \\
8 & 42.63 & 50 & 0.833 & 2.19\% & 44.96\% \\
16 & 59.21 & 72 & \textbf{1.156} & \textbf{81.49\%} &
\textbf{33.32\%} \\
\end{longtable}
}

The gap is roughly flat while \texttt{r\_bar\ \textless{}\ 1} and falls
sharply once the union routinely exceeds capacity. The measured unions
also fall below an independent uniform routing null by 3.1\% at B = 2
rising to 28.2\% at B = 16, so concurrent requests do overlap; the
crossover is nevertheless reached between B = 8 and B = 16.

\subsection{6.3 Same model, same workload, capacity
only}\label{same-model-same-workload-capacity-only}

Concurrency is not the only way to move \texttt{r\_bar}. Holding the
model, the trace and the concurrency fixed and varying only the
residency fraction produces the same effect.

\textbf{Table 10.} Gap versus residency fraction at fixed concurrency.
\emph{(128-expert model, per-layer scope, B = 16.)}

{\def\LTcaptype{none} 
\begin{longtable}[]{@{}rrrr@{}}
\toprule\noalign{}
\ensuremath{\rho} & best causal & Belady & recoverable gap \\
\midrule\noalign{}
\endhead
\bottomrule\noalign{}
\endlastfoot
20\% & 30.33\% & 28.08\% & \textbf{7.42\%} \\
30\% & 22.24\% & 18.09\% & 18.65\% \\
40\% & 15.58\% & 10.39\% & \textbf{33.32\%} \\
\end{longtable}
}

A study reporting only the \ensuremath{\rho} = 20\% row would conclude
that existing policies are within 7.4\% of optimal and that the problem
is closed. A study reporting only the \ensuremath{\rho} = 40\% row would
conclude that a third of the traffic is recoverable. Both are the same
model, the same workload and the same concurrency.

\subsection{\texorpdfstring{6.4 Cross-model comparison fails on batch
size and improves on
\texttt{r\_bar}}{6.4 Cross-model comparison fails on batch size and improves on r\_bar}}\label{cross-model-comparison-fails-on-batch-size-and-improves-on-r_bar}

Table 11 compares three models two ways: aligned on batch size, as is
conventional, and aligned on mean \texttt{r\_bar}.

\textbf{Table 11.} Recoverable gap under two alignment choices.
\emph{(\ensuremath{\rho} = 40\%, per-layer scope, event-atomic.)}

{\def\LTcaptype{none} 
\begin{longtable}[]{@{}llrrrr@{}}
\toprule\noalign{}
alignment & model & B & \texttt{r\_bar} & gap & spread \\
\midrule\noalign{}
\endhead
\bottomrule\noalign{}
\endlastfoot
\textbf{on batch size} & 40-expert & 8 & 1.75 & 8.17\% & \\
& 64-expert & 8 & 1.44 & 28.26\% & \\
& 128-expert & 8 & 0.83 & 44.96\% & \textbf{36.8 pp} \\
\textbf{on \texttt{r\_bar\ \textasciitilde{}=\ 0.30}} & 64-expert & 1 &
0.313 & 48.13\% & \\
& 128-expert & 2 & 0.293 & 49.25\% & \textbf{1.1 pp} \\
\textbf{on \texttt{r\_bar\ \textasciitilde{}=\ 0.51}} & 40-expert & 1 &
0.500 & 43.48\% & \\
& 128-expert & 4 & 0.519 & 48.12\% & \textbf{4.6 pp} \\
\end{longtable}
}

Aligned on batch size, the three models appear to behave qualitatively
differently --- one suggests the problem is essentially closed, another
that it is wide open. Aligned on \texttt{r\_bar}, the same measurements
agree to within a few percentage points. The apparent architecture
dependence was a regime difference.

This also bounds what small models can be used for. The 40-expert model
at \ensuremath{\rho} = 40\% cannot reach
\texttt{r\_bar\ \textless{}\ 0.5} at any concurrency, because a single
request already touches 8 of its 16 per-layer slots. Regimes below that
are simply not observable on it, and results from it cannot be
extrapolated into them.

\textbf{A note on the analytical reference frame.} For the
frontier-scale specification of §3.2 --- 896 experts, top-16,
\ensuremath{\rho} = 40\%, giving \texttt{c\_l\ \textasciitilde{}=\ 358}
--- the expected per-step union at B = 8 under a uniform routing null is
approximately 120, so \texttt{r\_bar\ \textasciitilde{}=\ 0.335} under
that null. That places it near the 64-expert model at B = 1 and the
128-expert model at B = 2, and far from any of them at B = 8. We state
this only to illustrate that the alignment matters at scale; it is an
analytical projection from published architecture parameters, not a
measurement.

\subsection{6.5 Necessary but not
sufficient}\label{necessary-but-not-sufficient}

\texttt{r\_bar} is a normalizer, not a sufficient statistic. To show
this directly we hold the multiset of event expert sets and
\texttt{r\_bar} \textbf{exactly} constant --- the same events, the same
sets, the same mean value 0.8237 --- and permute only their temporal
order.

The recoverable gap moves from 44.85\% to 30.79\%--31.06\%, and the best
causal policy changes from LFRU to LFU.

\begin{figure}
\centering
\includegraphics[width=0.78\linewidth,height=\textheight,keepaspectratio,alt={Figure 3. Matching mean union/cache is not sufficient. The original event order and all twenty permutations have the same event-set multiset and mean ratio (0.8237); only temporal order changes.}]{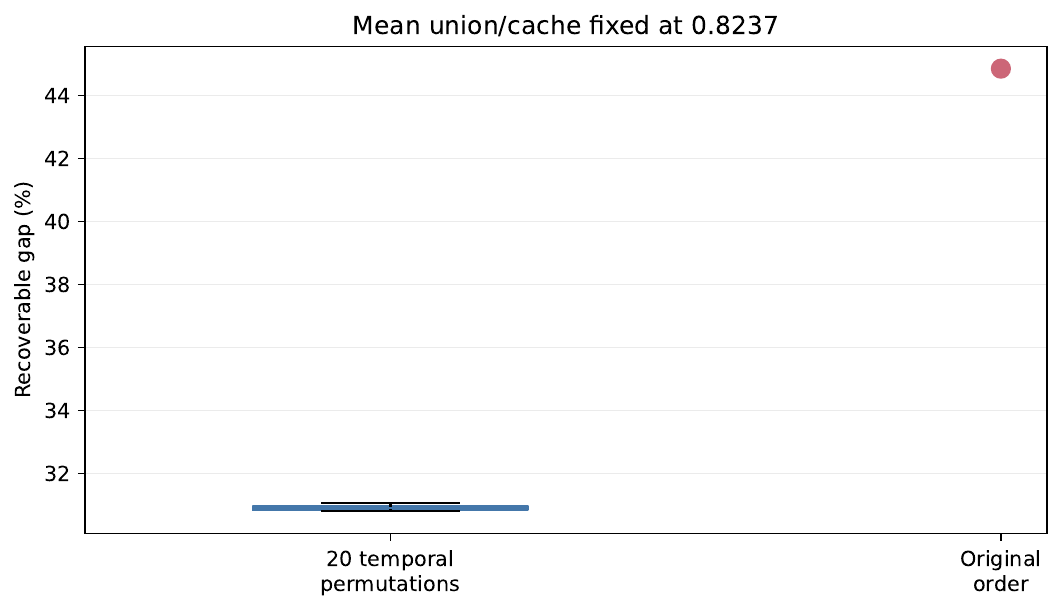}
\caption{\textbf{Figure 3. Matching mean union/cache is not sufficient.}
The original event order and all twenty permutations have the same
event-set multiset and mean ratio (0.8237); only temporal order
changes.}\label{fig:regime-insufficiency}
\end{figure}

The permutation does not represent a realizable request scheduler; it is
a mathematical sufficiency counterexample. It shows that reuse
structure, not just working-set size relative to capacity, determines
both the size of the gap and which policy realizes it. Quantities that
vary independently of \texttt{r\_bar} and that we expect to matter
include the inter-event drift of the working set, the reuse-distance
distribution, expert popularity concentration,
top-\texttt{k}/\texttt{N}, cache scope, and the scheduling discipline.
§9 gives an instance of the last: changing only the service discipline,
at fixed request set, capacity and mean batch size, moves a static
pinned set from 7.5\% worse than LFRU to 17.0\% better.

The correct prescription is therefore: \textbf{report the distribution
of \texttt{r(s,l)} and at least \texttt{r\_bar} --- without it,
cross-model comparison is not interpretable --- but do not treat
matching \texttt{r\_bar} as sufficient grounds for extrapolation.}

\subsection{6.6 Consequence: gap alone is not a decision
criterion}\label{consequence-gap-alone-is-not-a-decision-criterion}

Sections 6.2 and 6.3 have a practical corollary. A small gap is
ambiguous: it can mean that existing policies are close to optimal, or
that the system is in the thrashing regime where nothing helps. The two
are distinguished only by the absolute traffic relative to a transfer
budget.

Here ``offline traffic'' means Belady's transferred bytes under the
fixed access sequence, because causal traffic above budget does not by
itself establish physical infeasibility.

{\def\LTcaptype{none} 
\begin{longtable}[]{@{}
  >{\raggedright\arraybackslash}p{(\linewidth - 4\tabcolsep) * \real{0.3333}}
  >{\raggedright\arraybackslash}p{(\linewidth - 4\tabcolsep) * \real{0.3333}}
  >{\raggedright\arraybackslash}p{(\linewidth - 4\tabcolsep) * \real{0.3333}}@{}}
\toprule\noalign{}
\begin{minipage}[b]{\linewidth}\raggedright
\end{minipage} & \begin{minipage}[b]{\linewidth}\raggedright
small causal-to-offline gap
\end{minipage} & \begin{minipage}[b]{\linewidth}\raggedright
large causal-to-offline gap
\end{minipage} \\
\midrule\noalign{}
\endhead
\bottomrule\noalign{}
\endlastfoot
\textbf{offline traffic within budget} & existing policies suffice; a
controller is unnecessary & genuine algorithmic headroom on the fixed
sequence \\
\textbf{offline traffic over budget} & \textbf{no residency policy on
this sequence is viable} & oracle also fails; the gap is not enough to
cross the budget \\
\end{longtable}
}

We recommend that any reported gap be accompanied by the absolute
transferred bytes per output token and the bandwidth budget implied by
the target service rate. In our own use of this criterion, the smaller
models produced operating points in the lower-left cell --- gap under
10\% with the offline optimum still several times over budget --- that a
gap-only reading would have classified as ``existing policies already
suffice.''

We do not claim \texttt{r\ =\ 1} is a sharp threshold, only that
behaviour changes across it; the permutation ablation establishes
insufficiency rather than quantifying how much of the gap reuse
structure explains in general (§11).

\section{§7 The Offline-Optimal Gap and Its
Decomposition}\label{the-offline-optimal-gap-and-its-decomposition}

\begin{quote}
\textbf{Claim.} With all three axes controlled, a large gap to the
offline optimum remains and is stable across workload composition.
Decomposing it shows that 84.3--96.6\% derives from knowing which
resident block is used furthest in the future, a quantity for which the
natural online estimator not only fails to help but performs worse than
the baseline it replaces. A large offline-optimal gap is therefore not,
on its own, evidence of an online opportunity.
\end{quote}

\subsection{7.1 The gap is large and
stable}\label{the-gap-is-large-and-stable}

\textbf{Table 12.} Recoverable gap on the principal conditions.
\emph{(128-expert model, per-layer scope, \ensuremath{\rho} = 40\%,
event-atomic; best causal chosen from LRU, LFU, LFRU, Least-Stale.)}

{\def\LTcaptype{none} 
\begin{longtable}[]{@{}lrrr@{}}
\toprule\noalign{}
condition & best causal & Belady & gap \\
\midrule\noalign{}
\endhead
\bottomrule\noalign{}
\endlastfoot
held-out, B = 8 & 18.01\% (LFRU) & 9.93\% & 44.85\% \\
held-out, B = 2 & 12.84\% (LFRU) & 6.37\% & 50.42\% \\
calibration, B = 8 & 18.76\% (LFRU) & 10.35\% & 44.90\% \\
fixed-template arm, B = 8 & 21.99\% (LFRU) & 12.02\% & 45.33\% \\
\end{longtable}
}

The gap does not depend on the contamination of §5: the fixed-template
arm, whose absolute miss fraction is 22\% higher, shows the same gap to
within half a percentage point. Nor is it an artifact of a particular
workload mixture. Holding \ensuremath{\rho} = 40\%, B = 8 and the replay
semantics fixed, we constructed thirteen streams --- seven size-matched
mixtures drawn from disjoint request pools and six homogeneous
single-archetype streams --- and measured the gap on each.

\textbf{Table 13.} Gap invariance across workload composition. \emph{(13
deterministically frozen conditions, \ensuremath{\rho} = 40\%, B = 8,
per-layer scope. The conditions comprise six matched mixtures, one
non-matched mixture, and six non-matched pure streams.)}

{\def\LTcaptype{none} 
\begin{longtable}[]{@{}lr@{}}
\toprule\noalign{}
& gap \\
\midrule\noalign{}
\endhead
\bottomrule\noalign{}
\endlastfoot
minimum & 44.18\% \\
maximum & 45.93\% \\
mean & 45.09\% \\
\end{longtable}
}

The best causal policy is LFRU in twelve of thirteen conditions and LFU
in the homogeneous office/legal stream. \textbf{This range applies to
\ensuremath{\rho} = 40\%, B = 8 only.} It must not be pooled with the B
= 2 condition of Table 12 or with the residency sweep of §6.3, whose
gaps span 1.95\%--44.78\%; those are different operating points and
combining them would misrepresent the variability.

Across the four principal conditions, varying the tie-breaking seed
moves the gap by order \ensuremath{10^{-5}}. We report this as
implementation stability, not as a population confidence interval: it
bounds the simulator's nondeterminism, not the sampling variability of
the underlying traffic.

\subsection{7.2 What the offline optimum's advantage consists
of}\label{what-the-offline-optimums-advantage-consists-of}

Belady's advantage over a causal policy has two distinguishable sources.

\emph{Bypass admission.} On a miss, the optimum may decline to cache the
incoming block at all, if that block's next use is farther away than
that of every resident candidate. This has familiar causal
approximations in storage caching, including admission filters and reuse
prediction, but the oracle share measured below is not thereby
guaranteed to be recoverable.

\emph{Future-victim selection.} When the optimum does admit, it evicts
the resident block whose next use is farthest away. This asks for a
\emph{ranking} of all resident blocks by next-use distance, a
substantially harder estimation problem.

Both use future information; they differ in how tractable their online
analogues are. To separate them we run a third variant,
\texttt{Belady-forced-admit}, which retains furthest-next-use eviction
but is required to admit every miss. It is not a deployable policy ---
it still consults the future --- but it isolates the contribution of the
admission decision. Writing \texttt{m\_c}, \texttt{m\_B}, \texttt{m\_F}
for the miss fractions of the best causal policy, Belady, and
forced-admit Belady:

\begin{quote}
\texttt{admission\ share\ \ \ \ \ \ =\ (m\_F\ -\ m\_B)\ /\ (m\_c\ -\ m\_B)}
\texttt{future-victim\ share\ \ =\ (m\_c\ -\ m\_F)\ /\ (m\_c\ -\ m\_B)}
\end{quote}

\textbf{Table 14.} Decomposition of the recoverable gap.
\emph{(128-expert model, per-layer scope, \ensuremath{\rho} = 40\%,
held-out split; discovery values in parentheses.)}

{\def\LTcaptype{none} 
\begin{longtable}[]{@{}lrrr@{}}
\toprule\noalign{}
condition & total gap & admission share & future-victim share \\
\midrule\noalign{}
\endhead
\bottomrule\noalign{}
\endlastfoot
B = 8 & 44.85\% & 15.69\% (16.33\%) & \textbf{84.31\%} \\
B = 2 & 50.42\% & 3.40\% (3.37\%) & \textbf{96.60\%} \\
\end{longtable}
}

\begin{figure}
\centering
\includegraphics[width=0.72\linewidth,height=\textheight,keepaspectratio,alt={Figure 4. Most of the offline-optimal gap is future-victim knowledge in the two held-out operating points. Shares use event-atomic replay and the forced-admission decomposition; neither component is thereby guaranteed to be causally recoverable.}]{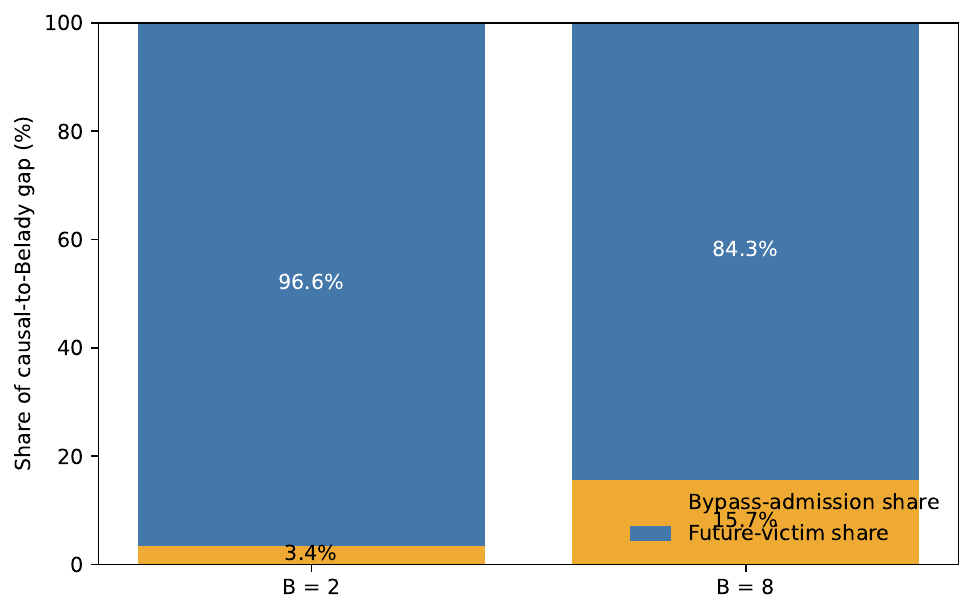}
\caption{\textbf{Figure 4. Most of the offline-optimal gap is
future-victim knowledge in the two held-out operating points.} Shares
use event-atomic replay and the forced-admission decomposition; neither
component is thereby guaranteed to be causally
recoverable.}\label{fig:gap-decomposition}
\end{figure}

The component with familiar online analogues is the minority in both
conditions, and it shrinks as concurrency falls. At B = 2 --- the
operating point whose mean regime ratio \texttt{r\_bar} is closest to
the frontier-scale reference frame of §6.4 --- 96.6\% of the gap rests
on ranking resident blocks by future distance.

\subsection{7.3 Can the dominant component be estimated
online?}\label{can-the-dominant-component-be-estimated-online}

The decomposition locates the gap; it does not establish that the
located quantity is unpredictable. We tested that directly by replacing
the oracle's input with a prediction and leaving everything else
unchanged.

We train a next-use-distance predictor on features available at the
moment of each access: log recency, a first-observation indicator, log
frequency, the log of the previous inter-access gap, the gate mass of
the access, the log of the number of concurrent requests routing to that
block in the same event, the layer index, and a popularity rate. The
last three are not used by any of our causal baselines, so the predictor
has strictly more information than LFRU. It is fitted on one trace and
evaluated on a disjoint one, and its output is substituted for the true
next-use array in the \emph{same} eviction and admission machinery,
yielding an algorithmically causal candidate. We do not measure its
inference cost or establish that it meets a production scheduling
budget.

\textbf{Table 15.} Predicted-next-use policy against the baselines it
replaces. \emph{(128-expert model, fitted on calibration split,
evaluated on held-out split, per-layer scope, \(\rho\) = 40\%, B = 8.)}

{\def\LTcaptype{none} 
\begin{longtable}[]{@{}lr@{}}
\toprule\noalign{}
& effective miss fraction \\
\midrule\noalign{}
\endhead
\bottomrule\noalign{}
\endlastfoot
LFU & 19.14\% \\
\textbf{LFRU (best causal)} & \textbf{18.01\%} \\
\textbf{predicted next-use} & \textbf{18.93\%} \\
Belady & 9.93\% \\
\textbf{fraction of the gap recovered} & \textbf{\(-11.39\)\%} \\
\end{longtable}
}

The predictor is informative as a regressor --- its transfer \(R^2\) on
log next-use distance is 0.237 --- and still makes the policy worse than
the baseline it was intended to improve. Regression accuracy over all
accesses is, however, the wrong measure: eviction needs a correct
\emph{ordering} among the blocks that are resident at the moment of
decision. We therefore instrumented the policy's eviction decisions
directly.

\textbf{Table 16.} Victim-ranking quality at eviction decision points,
tie-aware. \emph{(Same configuration; 1,708,374 eviction decisions,
46,172 sampled every 37th decision; mean candidate set 51.2 blocks; mean
optimal-victim set 1.24 blocks; 95\% CIs by cluster bootstrap over 3,055
scheduler steps.)}

{\def\LTcaptype{none} 
\begin{longtable}[]{@{}lrr@{}}
\toprule\noalign{}
victim chosen by & in the optimal-victim set & distance regret \\
\midrule\noalign{}
\endhead
\bottomrule\noalign{}
\endlastfoot
random candidate & 2.42\% {[}2.28, 2.56{]} & 0.00530 \\
\textbf{predicted next use} & \textbf{3.39\% {[}3.22, 3.57{]}} &
0.00519 \\
LRU & 20.64\% {[}20.20, 21.07{]} & 0.00315 \\
LFRU & \textbf{22.11\% {[}21.68, 22.56{]}} & 0.00301 \\
\end{longtable}
}

{\def\LTcaptype{none} 
\begin{longtable}[]{@{}
  >{\raggedright\arraybackslash}p{(\linewidth - 2\tabcolsep) * \real{0.4286}}
  >{\raggedleft\arraybackslash}p{(\linewidth - 2\tabcolsep) * \real{0.5714}}@{}}
\toprule\noalign{}
\begin{minipage}[b]{\linewidth}\raggedright
rank correlation, predicted vs.~true next use
\end{minipage} & \begin{minipage}[b]{\linewidth}\raggedleft
Spearman
\end{minipage} \\
\midrule\noalign{}
\endhead
\bottomrule\noalign{}
\endlastfoot
all resident candidates (average ranks for ties) & \(-0.207\)
{[}\(-0.210\), \(-0.204\){]} \\
excluding candidates never used again & \(-0.206\) {[}\(-0.209\),
\(-0.203\){]} \\
\end{longtable}
}

Because the optimal victim need not be unique, we score a choice as
correct if it attains the maximum true next-use key, not if it equals a
particular tie-broken identifier; Spearman uses average ranks. Ties are
in fact rare here --- the mean optimal set holds 1.24 blocks and on
average 0.07 candidates are never used again --- so neither convention
drives the result.

Three readings agree. The predicted ordering is \emph{anti-correlated}
with the truth, and this survives both tie-aware ranking and dropping
never-again candidates. The predictor selects an optimal victim 3.39\%
of the time, barely above the 2.42\% obtained by choosing a resident
block uniformly at random, and roughly six times less often than LRU or
LFRU at the identical cache state. Its distance regret is statistically
close to random and about 70\% worse than either causal baseline. The
predictor therefore supplies no usable ordering signal where the
decision is taken, which is what the end-to-end result of Table 15
reflects; because the same predictions also drive bypass admission,
errors compound, as a block whose next use is over-estimated is refused
admission and then immediately required.

A positive global \(R^2\) and a negative conditional rank correlation
can coexist. Selection induced by the policy is one possible explanation
--- the resident set is not a random sample of blocks, since survivors
are those the predictor scored as soon to be reused --- but we have not
isolated this experimentally, and we offer it as a hypothesis rather
than a finding.

\subsection{7.4 What this establishes, and what it does
not}\label{what-this-establishes-and-what-it-does-not}

We separate three claims and do not permit inference between them.

{\def\LTcaptype{none} 
\begin{longtable}[]{@{}
  >{\raggedright\arraybackslash}p{(\linewidth - 4\tabcolsep) * \real{0.3333}}
  >{\raggedright\arraybackslash}p{(\linewidth - 4\tabcolsep) * \real{0.3333}}
  >{\raggedright\arraybackslash}p{(\linewidth - 4\tabcolsep) * \real{0.3333}}@{}}
\toprule\noalign{}
\begin{minipage}[b]{\linewidth}\raggedright
\end{minipage} & \begin{minipage}[b]{\linewidth}\raggedright
claim
\end{minipage} & \begin{minipage}[b]{\linewidth}\raggedright
status
\end{minipage} \\
\midrule\noalign{}
\endhead
\bottomrule\noalign{}
\endlastfoot
1 & The full gap is the offline bound for residency policies on a fixed
access sequence & definitional \\
2 & Bypass admission contributes 3.4--15.7\%; future-victim selection
contributes 84.3--96.6\% & measured, stable across splits \\
3 & The causal approximation of §7.3 fails to recover the gap & specific
to these features, this predictor, this model, this workload, this
operating point \\
\end{longtable}
}

Claim 3 does not imply that future-victim distance is unpredictable in
principle, and we make no such claim. A predictor with materially higher
transfer accuracy, or one using features our traces do not contain,
could change the result.

It is also worth separating our negative result from a body of work that
is sometimes read as adjacent. Expert \emph{prefetching} predicts
\textbf{which experts a future step will route to}; our predictor
estimates \textbf{how long a cached block will remain unused}. These are
different targets with different horizons and different error costs, and
a positive result on the former is not evidence against the negative
result here (\citeproc{ref-fang2025fate}{Fang et al. 2025};
\citeproc{ref-du2024sidamoe}{Du et al. 2024};
\citeproc{ref-gavhane2025moebeyond}{Gavhane et al. 2025};
\citeproc{ref-zhu2026dali}{Zhu et al. 2026}).

What we do claim is narrower and, we believe, portable:

\begin{quote}
\textbf{In our evaluated settings a large offline-optimal gap
substantially overstates the gains recovered by representative
lightweight causal mechanisms.}
\end{quote}

The practical implication is a reporting requirement rather than a
design conclusion. A gap reported without decomposition invites the
reader to treat it as available headroom. We recommend that any work
motivating a cache controller by an offline-optimal gap also report the
forced-admission decomposition, so that the share of the gap resting on
future-victim knowledge is visible.

The gap itself is regime-dependent (§6), so these values are not
properties of MoE routing in general; the predictor is one model class
and no architecture search was performed (§11).

\section{§8 Scoped Negative Results and One Invalidated
Design}\label{scoped-negative-results-and-one-invalidated-design}

\begin{quote}
\textbf{Framing.} §7 shows that the offline-optimal gap is dominated by
a component whose natural online estimator does not help. This section
reports what happened when we pursued four mechanisms that do not
require estimating that component. Three produced scoped negative
results; the hard-partition study had no power to test its intended
mechanism because its dedicated cache was smaller than one request's
top-k set. We report each with the conditions under which it was
measured, because at least one fails for a reason that does not
generalize, and we do not claim mechanism-level closure.
\end{quote}

\subsection{8.1 The four attempted
mechanisms}\label{the-four-attempted-mechanisms}

Each decision threshold below was frozen, together with the split it
would be evaluated on, before the corresponding measurement was read.
The later failure attribution and the determination that the partition
grid was arithmetically incapable of a fair positive test are post-hoc
audits, not confirmatory results.

\textbf{Table 17.} Pre-registered criteria and outcomes, with
measurement scope.

{\def\LTcaptype{none} 
\begin{longtable}[]{@{}
  >{\raggedright\arraybackslash}p{(\linewidth - 6\tabcolsep) * \real{0.2500}}
  >{\raggedright\arraybackslash}p{(\linewidth - 6\tabcolsep) * \real{0.2500}}
  >{\raggedright\arraybackslash}p{(\linewidth - 6\tabcolsep) * \real{0.2500}}
  >{\raggedright\arraybackslash}p{(\linewidth - 6\tabcolsep) * \real{0.2500}}@{}}
\toprule\noalign{}
\begin{minipage}[b]{\linewidth}\raggedright
mechanism
\end{minipage} & \begin{minipage}[b]{\linewidth}\raggedright
frozen measurement setting
\end{minipage} & \begin{minipage}[b]{\linewidth}\raggedright
pre-registered criterion
\end{minipage} & \begin{minipage}[b]{\linewidth}\raggedright
measured outcome in scope
\end{minipage} \\
\midrule\noalign{}
\endhead
\bottomrule\noalign{}
\endlastfoot
steady-state prefill eviction / admission & 40- and 64-expert models;
\ensuremath{\rho} = 20/30/40\%; 128-token prefill blocks; strictly
lossless & gap \ensuremath{\geq} 10\% & 0.155--3.169\%; \textbf{not
met} \\
warm-cache workload transition & 40- and 64-expert models;
\ensuremath{\rho} = 20/30/40\%; directed A\ensuremath{\rightarrow}B
transitions; windows N = 10/25/50/100 & gap \ensuremath{\geq} 10\% &
1439/1440 units below 10\%; \textbf{not met} \\
semantic category partitioning & 128-expert model; \ensuremath{\rho} =
40\%; B = 8; FCFS; equal total capacity; \textbf{6 slots \textless{}
top-\texttt{k} = 8} & transfer reduction \ensuremath{\geq} 5\% &
transfer increased 115.9\%; \textbf{invalid test of the mechanism
because the design floor dominated} \\
causal affinity batching & 128-expert model; \ensuremath{\rho} = 40\%; B
= 8 with 24 admitted; deadline rotation; W = 4; strictly lossless &
transfer reduction \ensuremath{\geq} 10\% & 4.76\% (LFRU) / 4.55\%
(static); \textbf{not met} \\
\end{longtable}
}

The warm-transition experiment covers 1440 units --- two models
\ensuremath{\times} three residency fractions \ensuremath{\times} three
request-order seeds \ensuremath{\times} twenty directed transitions
between five workload proxies \ensuremath{\times} four post-transition
windows. A single unit exceeded the threshold, at 10.74\%, and decayed
to 6.83\%, 5.03\% and 3.95\% as the window widened from 10 to 25, 50 and
100 requests, which is the signature of a short-lived warm-start
transient rather than a sustainable control opportunity.

\subsection{8.2 One invalidated design}\label{one-invalidated-design}

The partitioning experiment gave a dedicated per-layer slot budget to
each of the two archetypes §5.7 identifies as genuinely cache-friendly,
with the remaining four sharing a third partition, at exactly equal
total capacity. Every allocation in a seven-point grid lost, and the
frozen one increased transfer by 115.9\%.

The natural reading --- that partitioning destroys cross-category expert
sharing --- is largely wrong. Running the \emph{shared} cache at each
partition's own slot count and weighting by each partition's share of
logical assignments predicts 37.30\% against a measured 37.99\%:
\textbf{capacity fragmentation alone accounts for 96\% of the
degradation.} The cause is arithmetic. Top-\texttt{k} is 8, so a single
request touches eight experts per layer, while the best dedicated
partition holds six; no point in the grid could retain even one
request's per-step working set. \textbf{The experiment as designed could
not have produced a positive result}, and we present it as an
invalidated design and a measurement lesson (§10, item D5) rather than
as a negative result for semantic partitioning. The arithmetic does not
generalize: at the reference scale of §3.2, 358 per-layer slots divided
six ways leave a 2.3--3.7x margin over the per-category per-step union,
so the \(-115.9\)\% figure must not be carried across. Appendix B gives
the grid.

\subsection{8.3 Affinity batching: not a prediction
problem}\label{affinity-batching-not-a-prediction-problem}

The remaining mechanism does not touch residency at all. It changes
which requests are co-scheduled, and therefore changes the event
sequence itself --- a degree of freedom that the Belady bound of §7 does
not cover, since that bound is defined on a fixed access sequence.

Up to 24 sessions are admitted and at most \texttt{B\ =\ 8} are served
per step, with a frozen fairness constraint that an admitted request's
inter-service interval never exceeds \texttt{W\ =\ 4} steps. Three
schedulers share identical arrival, admission and completion rules and
differ only in how they select up to eight requests from the active set:
\texttt{fcfs\_deadline}; \texttt{causal\_prev\_route}, which uses only
each request's per-layer expert set at its previous executed token and
greedily minimizes the predicted union; and
\texttt{oracle\_current\_route}, which runs the same selection using the
true expert sets of the tokens about to execute. The oracle is a
rolling, route-aware bound, not a global optimum, and all metrics are
computed on the complete cache trajectory rather than per step.

\textbf{Table 18.} Causal affinity batching against FCFS, held-out
split. \emph{(128-expert model, \ensuremath{\rho} = 40\%, per-layer
scope, B = 8, 24 admitted, W = 4.)}

{\def\LTcaptype{none} 
\begin{longtable}[]{@{}
  >{\raggedright\arraybackslash}p{(\linewidth - 14\tabcolsep) * \real{0.0968}}
  >{\raggedleft\arraybackslash}p{(\linewidth - 14\tabcolsep) * \real{0.1290}}
  >{\raggedleft\arraybackslash}p{(\linewidth - 14\tabcolsep) * \real{0.1290}}
  >{\raggedleft\arraybackslash}p{(\linewidth - 14\tabcolsep) * \real{0.1290}}
  >{\raggedleft\arraybackslash}p{(\linewidth - 14\tabcolsep) * \real{0.1290}}
  >{\raggedleft\arraybackslash}p{(\linewidth - 14\tabcolsep) * \real{0.1290}}
  >{\raggedleft\arraybackslash}p{(\linewidth - 14\tabcolsep) * \real{0.1290}}
  >{\raggedleft\arraybackslash}p{(\linewidth - 14\tabcolsep) * \real{0.1290}}@{}}
\toprule\noalign{}
\begin{minipage}[b]{\linewidth}\raggedright
policy
\end{minipage} & \begin{minipage}[b]{\linewidth}\raggedleft
FCFS
\end{minipage} & \begin{minipage}[b]{\linewidth}\raggedleft
causal
\end{minipage} & \begin{minipage}[b]{\linewidth}\raggedleft
oracle
\end{minipage} & \begin{minipage}[b]{\linewidth}\raggedleft
causal gain
\end{minipage} & \begin{minipage}[b]{\linewidth}\raggedleft
oracle gain
\end{minipage} & \begin{minipage}[b]{\linewidth}\raggedleft
share
\end{minipage} & \begin{minipage}[b]{\linewidth}\raggedleft
\ensuremath{\Delta}p99
\end{minipage} \\
\midrule\noalign{}
\endhead
\bottomrule\noalign{}
\endlastfoot
LFRU & 23.08\% & 21.98\% & 21.71\% & \textbf{4.76\%} & 5.92\% &
\textbf{80.43\%} & +7.06\% \\
static pinned & 19.15\% & 18.28\% & 17.64\% & \textbf{4.55\%} & 7.90\% &
57.65\% & +10.08\% \\
\end{longtable}
}

No request starved and the maximum inter-service interval was 4
throughout. The static-residency arm additionally violated the frozen
p99 constraint at 10.08\%.

The important number is the second-to-last column. The causal scheduler
captures 80.4\% of what the route-aware oracle achieves under the same
fairness constraint. \textbf{The shortfall is therefore not a weakness
of the prediction signal}; it is that the schedulable headroom under
\texttt{W\ =\ 4} at this concurrency is itself only 5.9--7.9\%.
Improving the predictor cannot reach a 10\% threshold that the oracle
does not reach either. This distinguishes the result from §7.3, where
the causal estimator was the binding constraint.

The signal itself is real: within a request, the per-layer expert set at
step \texttt{t+1} overlaps that at step \texttt{t} with mean Jaccard
0.293, against 0.032 for two independent uniform draws --- a ninefold
enrichment. It is simply not worth much once fairness is enforced.

\subsection{8.4 A methodological consequence: single-step order
statistics}\label{a-methodological-consequence-single-step-order-statistics}

Before running §8.3 we estimated the headroom the cheap way, by sampling
300 random batch compositions at a fixed step and comparing the best to
the mean union. That estimate was \textbf{19.3\%}. The full-trajectory
measurement under the same concurrency with \texttt{W\ =\ 4} yields an
oracle headroom of \textbf{5.9--7.9\%}.

Two effects account for the difference and both are general. The maximum
over 300 draws is an order statistic and is biased upward as an estimate
of what a systematic method achieves. More importantly, a per-step
measurement ignores conservation: requests that are deferred because
they route poorly with the current batch must still be served later, and
their cost reappears. A per-step union reduction is therefore an upper
bound that can be almost entirely recovered by the later schedule.

We recommend that scheduling headroom for MoE caching be estimated on
complete trajectories under the intended fairness constraint, and that
per-step best-of-\texttt{N} figures not be reported as achievable gains.

\subsection{8.5 What these results do not
close}\label{what-these-results-do-not-close}

We state the complement explicitly, because the value of a negative
result depends on its boundary.

\begin{itemize}
\tightlist
\item
  \textbf{Prefill execution ordering.} §8.1 finds little
  eviction/admission headroom in the measured prefill conditions only.
  Layer-major versus chunk-major prefill scheduling was never measured;
  under an independent uniform-routing approximation the two differ by
  roughly two orders of magnitude in total expert traffic for long
  contexts, and it remains the largest unexamined lever we are aware of.
\item
  \textbf{Lossy settings.} All results assume strictly lossless service.
  Expert substitution, pruning, and reduced-precision fallback are
  excluded by construction.
\item
  \textbf{Prefetching.} We evaluate residency and scheduling, not
  prefetch. §7.4 states why our negative result does not bear on expert
  prediction for prefetching.
\item
  \textbf{Frontier scale.} Every mechanism was measured at
  \texttt{r\_bar} values reachable by models of 40--128 experts. §6.4
  shows that the reference frame of §3.2 sits at a mean regime those
  models reach only at low concurrency.
\item
  \textbf{Other fairness settings.} §8.3 fixes \texttt{W\ =\ 4} and 24
  admitted sessions; relaxing either enlarges the schedulable headroom
  and was not swept.
\end{itemize}

\subsection{8.6 Recorded design
revision}\label{recorded-design-revision}

One parameter in §8.3 was revised before execution. The initial draft
admitted \texttt{B\ x\ W\ =\ 32} sessions; a pre-execution check showed
this implies exactly 100\% service load, leaving no discretion for
affinity selection to exercise. The count was reduced to 24, and the
revision was recorded together with the statement that no result had
been generated or read, and with an explicit prohibition on relaxing
\texttt{W} or reverting the count afterwards. The document is released
unchanged.

\section{§9 What Survives}\label{what-survives}

Three corrections, three scoped negative results, and one invalidated
design leave a small set of descriptive statements and baselines
statements. This section reports them, together with one confounder we
did not anticipate and found only because a mechanism experiment
happened to use a different service discipline.

\subsection{9.1 A static pinned set is a simple reference, not a
frontier}\label{a-static-pinned-set-is-a-simple-reference-not-a-frontier}

The simplest possible residency policy holds a fixed set of experts
chosen offline from a calibration trace and never updates it. It
requires no counters, no eviction logic, and no runtime state. Table 19
compares it against dynamic policies under standard continuous batching,
using a pin list fitted on the calibration split and frozen before the
held-out split was read.

\textbf{Table 19.} Static pinned residency versus dynamic policies
across concurrency. \emph{(128-expert model, held-out split, per-layer
scope, \ensuremath{\rho} = 40\%, event-atomic; pin list frozen from the
calibration split; initial load counted.)}

{\def\LTcaptype{none} 
\begin{longtable}[]{@{}rrrrr@{}}
\toprule\noalign{}
active B & static pinned & LFRU & Belady & static vs LFRU \\
\midrule\noalign{}
\endhead
\bottomrule\noalign{}
\endlastfoot
2 & 21.50\% & 12.84\% & 6.37\% & \textbf{\textminus{}67.4\%} \\
4 & 20.81\% & 16.26\% & 8.43\% & \textminus{}28.0\% \\
8 & 19.37\% & 18.01\% & 9.93\% & \textminus{}7.5\% \\
16 & 17.10\% & 16.64\% & 11.32\% & \textminus{}2.8\% \\
24 & 15.28\% & 14.92\% & 12.21\% & \textminus{}2.4\% \\
32 & 13.84\% & 13.52\% & 12.20\% & \textbf{\textminus{}2.3\%} \\
\end{longtable}
}

The static set never wins, but the margin narrows monotonically from a
factor of 1.7 at B = 2 to 2.3\% at B = 32. The mechanism is the regime
variable of §6: at low concurrency the per-step working set is small
relative to capacity and a dynamic policy can track it precisely,
whereas a fixed set must cover the global popularity distribution; as
concurrency grows the union approaches and then exceeds capacity and the
advantage of tracking disappears.

At B=8 and above the pin list transfers with a modest absolute penalty,
but at B=2 it is 67.4\% worse than LFRU. This matters because a
frontier-scale model at the same residency fraction may occupy a
similarly slack union/capacity regime; our local data therefore do not
support a product claim for static pinning.

The B=8 protocol audit makes the fit-source comparison explicit. On the
same held-out event stream and with identical equal per-layer quotas,
\texttt{Static-frozen} transfers 19.3684\% of logical assignments, while
the leaky \texttt{Static-same-trace} diagnostic transfers 19.1004\%. The
frozen list therefore pays 0.268 percentage points, or 1.403\% relative,
on this one held-out stream. Both numbers count the initial 2,458-block
preload. We make no claim about an optimized nonuniform layer
allocation: an earlier draft quoted such a result without a frozen
artifact, so it is removed rather than reconstructed after the fact.

\subsection{9.2 Static residency performance is a function of the
popularity distribution
alone}\label{static-residency-performance-is-a-function-of-the-popularity-distribution-alone}

Because a static set never evicts, its steady-state miss count is simply
the number of accesses to non-resident blocks. Measured over the
held-out trace, the access distribution is only moderately concentrated:

{\def\LTcaptype{none} 
\begin{longtable}[]{@{}rr@{}}
\toprule\noalign{}
fraction of blocks & share of accesses \\
\midrule\noalign{}
\endhead
\bottomrule\noalign{}
\endlastfoot
top 10\% & 24.8\% \\
top 20\% & 44.0\% \\
top 30\% & 59.6\% \\
top 40\% & \textbf{72.3\%} \\
\end{longtable}
}

For the \texttt{Static-same-trace} diagnostic at \ensuremath{\rho} =
40\%, the identity predicts a miss ratio of 1 \textminus{} 0.723 =
27.99\% of \textbf{union accesses}, and the simulator reports 27.99\%
under that denominator before adding the one-time preload. This is not
the pre-dedup effective-miss fraction used in Table 19. The identity is
exact for the union-access denominator: static residency traffic depends
only on the access-count distribution and not on access order.

This has a useful consequence. Unlike every other policy in this paper,
the performance of a static pinned set can be computed from an
\textbf{aggregate per-layer, per-expert access histogram}, with no
sequential trace required. Predicting how a dynamic policy will behave
requires the sequence; predicting how a static one will behave does not.

\subsection{9.3 A fourth confounder: service
discipline}\label{a-fourth-confounder-service-discipline}

§8.3 evaluated affinity batching under a deadline-driven rotation in
which up to 24 sessions are admitted and eight are served per step. We
had also measured the same request set under standard continuous
batching. The two disagree about which residency policy is better.

\textbf{Table 20.} Static versus dynamic residency under two service
disciplines. \emph{(Identical 72 requests, 23,529 decode forwards,
capacity and pin list; 128-expert model, \ensuremath{\rho} = 40\%.)}

{\def\LTcaptype{none} 
\begin{longtable}[]{@{}
  >{\raggedright\arraybackslash}p{(\linewidth - 8\tabcolsep) * \real{0.1579}}
  >{\raggedleft\arraybackslash}p{(\linewidth - 8\tabcolsep) * \real{0.2105}}
  >{\raggedleft\arraybackslash}p{(\linewidth - 8\tabcolsep) * \real{0.2105}}
  >{\raggedleft\arraybackslash}p{(\linewidth - 8\tabcolsep) * \real{0.2105}}
  >{\raggedleft\arraybackslash}p{(\linewidth - 8\tabcolsep) * \real{0.2105}}@{}}
\toprule\noalign{}
\begin{minipage}[b]{\linewidth}\raggedright
service discipline
\end{minipage} & \begin{minipage}[b]{\linewidth}\raggedleft
mean batch
\end{minipage} & \begin{minipage}[b]{\linewidth}\raggedleft
static pinned
\end{minipage} & \begin{minipage}[b]{\linewidth}\raggedleft
LFRU
\end{minipage} & \begin{minipage}[b]{\linewidth}\raggedleft
static vs LFRU
\end{minipage} \\
\midrule\noalign{}
\endhead
\bottomrule\noalign{}
\endlastfoot
continuous batching, B = 8 & 7.69 & 19.37\% & 18.01\% &
\textbf{\textminus{}7.5\%} \\
deadline rotation, 24 admitted / 8 served, W = 4 & 7.83 & 19.15\% &
23.08\% & \textbf{+17.0\%} \\
\end{longtable}
}

The static arm barely moves (19.37\% \ensuremath{\rightarrow} 19.15\%);
LFRU degrades by five percentage points. Rotating a subset of admitted
sessions repeatedly disrupts the recency and frequency state on which
dynamic policies depend, while a fixed set has no state to disrupt.

The two disciplines are not equally representative. Production
continuous batching advances every admitted sequence each iteration up
to a concurrency limit; queued requests wait rather than being rotated
in, and rotation arises mainly under memory-pressure preemption. The
rotation used in §8.3 was chosen to create the scheduling slack affinity
selection requires, not because it models a default deployment.

We report this as a measurement result rather than a design
recommendation. It means that \textbf{a residency-policy comparison that
does not state its service discipline is not interpretable}, on the same
footing as the three axes of §4--§6. We did not anticipate it, and we
found it only because two experiments run for different purposes
happened to disagree.

\subsection{9.4 Gate-mass rank is descriptive, not a quality
result}\label{gate-mass-rank-is-descriptive-not-a-quality-result}

Every mechanism in §8 attempts to reduce the \emph{number} of misses. An
orthogonal approach reduces the \emph{bytes per miss}: serve low-weight
experts at reduced precision while keeping high-weight experts at full
precision. This changes traffic without changing hit rate, so it is
untouched by the Belady bound.

Its viability depends on how unevenly gate mass is distributed across
the top-\texttt{k} ranks. On the 128-expert model it is nearly flat:

{\def\LTcaptype{none} 
\begin{longtable}[]{@{}
  >{\raggedright\arraybackslash}p{(\linewidth - 16\tabcolsep) * \real{0.0857}}
  >{\raggedleft\arraybackslash}p{(\linewidth - 16\tabcolsep) * \real{0.1143}}
  >{\raggedleft\arraybackslash}p{(\linewidth - 16\tabcolsep) * \real{0.1143}}
  >{\raggedleft\arraybackslash}p{(\linewidth - 16\tabcolsep) * \real{0.1143}}
  >{\raggedleft\arraybackslash}p{(\linewidth - 16\tabcolsep) * \real{0.1143}}
  >{\raggedleft\arraybackslash}p{(\linewidth - 16\tabcolsep) * \real{0.1143}}
  >{\raggedleft\arraybackslash}p{(\linewidth - 16\tabcolsep) * \real{0.1143}}
  >{\raggedleft\arraybackslash}p{(\linewidth - 16\tabcolsep) * \real{0.1143}}
  >{\raggedleft\arraybackslash}p{(\linewidth - 16\tabcolsep) * \real{0.1143}}@{}}
\toprule\noalign{}
\begin{minipage}[b]{\linewidth}\raggedright
rank
\end{minipage} & \begin{minipage}[b]{\linewidth}\raggedleft
1
\end{minipage} & \begin{minipage}[b]{\linewidth}\raggedleft
2
\end{minipage} & \begin{minipage}[b]{\linewidth}\raggedleft
3
\end{minipage} & \begin{minipage}[b]{\linewidth}\raggedleft
4
\end{minipage} & \begin{minipage}[b]{\linewidth}\raggedleft
5
\end{minipage} & \begin{minipage}[b]{\linewidth}\raggedleft
6
\end{minipage} & \begin{minipage}[b]{\linewidth}\raggedleft
7
\end{minipage} & \begin{minipage}[b]{\linewidth}\raggedleft
8
\end{minipage} \\
\midrule\noalign{}
\endhead
\bottomrule\noalign{}
\endlastfoot
share of gate mass & 21.5\% & 16.4\% & 13.7\% & 11.9\% & 10.5\% & 9.4\%
& 8.6\% & 8.0\% \\
\end{longtable}
}

The highest-ranked expert carries only 2.7\ensuremath{\times} the mass
of the lowest, and ranks 1--4 account for 63.5\% of the total. This says
that rank is a weak byte-allocation heuristic on this model. It does
\textbf{not} measure the quality loss from reducing precision: gate mass
is not causal output importance, quantization error is not equivalent to
dropping a contribution, and no mixed-precision implementation was
evaluated. We therefore retain the distribution as a design warning, not
as evidence that mixed precision is closed or that a larger model will
be flatter.

A cross-model comparison at aligned regimes suggests the static
disadvantage may grow with expert count, but the evidence is two
comparable points on heterogeneously collected traces; we report it in
Appendix B as hypothesis-generating only.

\section{§10 Benchmarking
Recommendations}\label{benchmarking-recommendations}

Each of the preceding sections identifies an evaluation choice that is
rarely stated and that changes conclusions when it varies. We collect
them here as a reporting checklist. Every item is one we failed to
report, or reported incorrectly, at some point in this work; the section
reference gives the evidence that motivated it. We apply the checklist
to ourselves in §13.

The intent is comparability rather than compliance. Most items ask an
author to \emph{state} a choice, not to make a particular one. Two items
--- A2 and D3 --- are stronger, because the practices they exclude are
not alternative modelling choices but errors.

\subsection{10.1 Group A --- Replay and execution
model}\label{group-a-replay-and-execution-model}

\textbf{Applicability.} Group A applies to trace-driven or otherwise
modelled cache evaluation. For an end-to-end real-system experiment, the
implementation itself defines the transfer unit and retention order; A1
should report that contract, while A2--A4 may be marked not applicable
rather than imposed retroactively.

\textbf{A1. State the execution unit.} Is hit/miss classified once per
\texttt{(step,\ layer)} event over the batch-wide expert union, or per
individual expert access? The two are not interchangeable. \emph{(§2.1)}

\textbf{A2. State whether retention decisions are deferred to the event
boundary.} If they are not, state explicitly whether a block resident at
event start and required later can be evicted and counted again. Such a
replay is inconsistent with one-transfer-per-distinct-expert fused-event
accounting; it may instead describe a genuinely serialized execution,
which should be named as a different contract. \emph{(§2.2, §4)}

\textbf{A3. If replay is sequential, state the intra-event ordering
rule.} Ascending expert index makes checkpoint numbering part of the
retention decision; randomizing it removes that dependence but not the
error in A2. \emph{(§4.1--§4.2)}

\textbf{A4. Provide a small hand-verifiable trace or equivalent
conformance check}, so that a reader can confirm which semantics an
implementation realizes. \emph{(§2.5)}

\subsection{10.2 Group B --- Workload
construction}\label{group-b-workload-construction}

\textbf{B1. Report the number of distinct instruction templates per
workload category}, together with the number of distinct prompt prefixes
and distinct final lines per category. A single template per category is
a risk indicator, not a defect by itself. \emph{(§5.6)}

\textbf{B2. State whether the model's chat template was applied.} Under
raw continuation the model may complete the instruction rather than
answer it, producing shared prefixes across all requests in a category.
\emph{(§5.2)}

\textbf{B3. Report the number of decode steps recorded} and whether the
analysis window extends beyond the shared-prefix region. Short windows
are dominated by it. \emph{(§5.2)}

\textbf{B4. State whether concurrent requests are position-locked} ---
equal length, lock-step admission and retirement --- since this aligns
any shared prefix at identical scheduling steps and inflates per-step
union effects. \emph{(§5.2)}

\textbf{B5. Report positional token agreement and the number of unique
generated sequences per category.} \emph{(§5.3)}

\textbf{B6. Report within-category expert overlap restricted to request
pairs whose generated text barely overlaps, against a cross-category
baseline.} The residual is the portion of the effect not explained by
surface repetition. \emph{(§5.3)}

\textbf{B7. Report whether the effect is stable across decode-position
bands.} An effect concentrated in the same band as a shared generated
prefix should be treated as surface repetition until shown otherwise.
Position decay without a shared prefix can be a genuine response-phase
effect. \emph{(§5.4)}

\textbf{B8. Where feasible, include a matched-pair control} in which the
same source records are rendered under both constructions, which
converts the artifact from an argument into a measurement. \emph{(§5.5)}

\subsection{10.3 Group C --- Regime and
comparability}\label{group-c-regime-and-comparability}

\textbf{C1. Report the per-step, per-layer expert-union distribution,
the per-layer capacity, event-specific \texttt{r(s,l)}, and at least its
mean \texttt{r\_bar} and fraction above one.} Without these a result
cannot be placed; the mean alone is not a sufficient statistic.
\emph{(§6.1, §6.5)}

\textbf{C2. State the cache scope} --- per-layer or global --- and, for
per-layer, how the remainder of the budget is allocated. \emph{(§2.3)}

\textbf{C3. State the miss denominator}: expert assignments before or
after intra-event deduplication. \emph{(§2.4)}

\textbf{C4. State whether cross-request deduplication is credited to the
policy.} It is obtained by any batching scheduler and is not a policy
contribution. \emph{(§2.4)}

\textbf{C5. State the service discipline}: continuous batching, rotation
of a subset of admitted sessions, admission and preemption rules, and
any fairness bound. Changing only this, at fixed request set, capacity
and mean batch size, moved a static pinned set from 7.5\% worse than
LFRU to 17.0\% better. \emph{(§9.3)}

\textbf{C6. For every static or pinned baseline, state its fit source,
split boundary, quota rule, and whether the initial load is counted} in
reported traffic. A same-trace popularity diagnostic and a
discovery-frozen pin list are different baselines even if both are
called ``static''. \emph{(§2.3--§2.4, §9.1)}

\textbf{C7. For cross-model comparison, first align on \texttt{r\_bar},
not on batch size, and then report remaining architectural differences.}
Aligned on batch size, three models in this paper span 36.8 percentage
points of gap; aligned on \texttt{r\_bar}, the same measurements agree
to within 1.1--4.6. \emph{(§6.4)}

\subsection{10.4 Group D --- Interpreting oracle
bounds}\label{group-d-interpreting-oracle-bounds}

\textbf{D1. Report absolute transferred bytes per output token alongside
any gap}, together with the bandwidth budget implied by the target
service rate. A small gap is ambiguous between ``existing policies
suffice'' and ``no policy is viable'', and only the absolute figure
distinguishes them. \emph{(§6.6)}

\textbf{D2. When motivating a controller by an offline-optimal gap,
report the forced-admission decomposition}, so that the share of the gap
resting on future-victim ranking is visible rather than implied to be
available. \emph{(§7.2)}

\textbf{D3. Do not report per-step best-of-\texttt{N} batch compositions
as achievable scheduling gains.} Estimate scheduling headroom on
complete trajectories under the intended fairness constraint; the
per-step figure in this work overstated the trajectory figure by a
factor of roughly three. \emph{(§8.4)}

\textbf{D4. State the scope of every negative result}: model, residency
fraction, concurrency, service discipline, fairness constraint, and
whether service is strictly lossless. One attempted negative result in
this paper was invalidated by its own capacity floor and would be
misleading without this scope. \emph{(§8.2, §8.5)}

Appendix B gives a minimal reporting block, compact enough to include
verbatim in a paper or artifact README, covering the items above that
admit a one-line answer.

\subsection{10.6 What the checklist does not
cover}\label{what-the-checklist-does-not-cover}

The checklist addresses trace-driven evaluation of residency and
scheduling. It says nothing about several things that matter for a
deployment claim and that we did not measure: real host-to-device
transfer and its overlap with compute, interconnect topology and
achievable sustained bandwidth, kernel time, memory contention between
expert cache and attention state, and any quality-lossy mechanism. A
study that satisfies every item here has established comparability with
other trace-driven studies, not deployability.

We also expect the list to be incomplete. §9.3 documents a confounder we
did not anticipate and found by accident, after three others had already
been identified; we see no reason to believe it is the last.

\section{§11 Limitations}\label{limitations}

This paper argues that MoE cache results are sensitive to unreported
evaluation choices, which obliges us to state our own boundaries
plainly.

\textbf{Everything here is trace-driven simulation.} We replay recorded
routing decisions against a cache model and count blocks transferred. We
do not measure host-to-device transfer, its overlap with compute,
sustained bandwidth, interconnect topology, kernel time, graph-capture
interactions, or contention between the expert cache and attention
state. Miss bursts are reported in block counts, not in time. A policy
that transfers fewer blocks here may not be faster in a system, and one
that transfers more may be, if its transfers overlap better. No
deployment claim follows from any result in this paper; where
engineering units appear they are unit conversions applied to the
analytical reference frame of §3.2 and are labelled as such.

\textbf{Model and scale coverage is narrow.} The results of §5, §7, §8
and §9 rest on a single 128-expert model, which is additionally 4-bit
quantized; routing under quantization may differ from full precision and
we did not compare them. The 40- and 64-expert models appear only for
cross-model checks. No trace was collected at the 896-expert scale that
motivates the application, and §6.4 shows that its regime ratio is
reached by our models only at low concurrency --- so the setting the
motivation comes from is the one our data covers least well. Regime
coverage is also bounded below by expert count: the 40-expert model
cannot reach \texttt{r\_bar\ \textless{}\ 0.5} at \ensuremath{\rho} =
40\% at any concurrency, so conclusions about slack regimes rest on the
128-expert model alone.

\textbf{Workload and probe construction constrain what the conclusions
mean.} We evaluate strictly lossless service only; expert substitution,
pruning, merging and reduced-precision fallback are excluded by
construction, and §9.4 measures the gate-mass distribution that would
govern one such mechanism without evaluating any implementation of it.
§8.1 closes prefill eviction and admission only --- layer-major versus
chunk-major prefill execution ordering was never measured and remains
the largest lever this paper does not examine. All traces use greedy
decoding, which is reproducible but does not sample production
generation diversity. The probe set repackages public sources; its
category proportions are a design choice and estimate no deployment's
traffic mix, and the six archetypes are not a taxonomy of enterprise
workloads. §9.1 fits a static residency set from one calibration trace
and measures transfer to one held-out trace; we do not study how quickly
such a set becomes stale under workload drift, which is the question a
deployment would ask.

\textbf{Method-specific and statistical limits.} The next-use estimator
of §7.3 is a linear model over eight features with no architecture
search; a negative result from a single estimator is weaker than the
decomposition it accompanies, and we report it because its failure is
not marginal and because it is what a practitioner would try first. The
forced-admission decomposition is arithmetic given a trace and inherits
every limitation above. §9.3 compares two service disciplines, not a
family, and §8.3 fixes one fairness setting (\texttt{W\ =\ 4}, 24
admitted sessions), so its 5.9--7.9\% oracle headroom is specific to
that constraint. Tie-seed variation of order \(10^{-5}\) bounds
simulator nondeterminism, not sampling variability; we report no
population confidence intervals for quantities estimated from a single
trace. Where draw-to-draw spread is quantified (§5.7) it is an estimate
of reference variability from seven size-matched draws, and each
homogeneous condition is still two draws of twelve requests. Reported
ranges apply to one operating point: the 44.18--45.93\% range of §7.1
covers thirteen workload compositions at \ensuremath{\rho} = 40\%, B =
8, and must not be pooled with the B = 2 condition or the residency
sweep, whose gaps span 1.95--44.78\%. Discovery and confirmatory splits
share no source group but were collected under the same model, decoding
configuration and length cap. Finally, the audit in §12 evaluates what
each work \textbf{reports}, not what it does; it establishes that
certain results cannot be compared, not that any is incorrect.

\textbf{The two that matter most.} If one limitation were removed we
would choose real-hardware validation: every result here is a block
count, and the step to service quality passes through overlap, topology
and contention we do not model. If a second were removed we would choose
a routing trace at the scale the motivation comes from --- because §6
establishes that these results are regime-dependent, the measurement
that would most change our confidence is not a better policy or a
stronger predictor but a trace from that regime.

\section{§12 Related Work and Reporting
Audit}\label{related-work-and-reporting-audit}

\subsection{12.1 Expert caching and offloading
systems}\label{expert-caching-and-offloading-systems}

Recent work attacks MoE memory pressure through expert caching,
prefetching, mixed-precision fallback, expert substitution, and CPU--GPU
co-execution. Mixtral-Offloading
(\citeproc{ref-eliseev2023mixtraloffloading}{Eliseev and Mazur 2023}),
MoE-Infinity (\citeproc{ref-xue2025moeinfinity}{Xue et al. 2025}), and
HOBBIT (\citeproc{ref-tang2024hobbit}{Tang et al. 2024}) implement
end-to-end offloading systems under different hardware and precision
assumptions. SpecMD (\citeproc{ref-hoang2026specmd}{Hoang et al. 2026})
provides a PyTorch-hook framework for composing caching, prefetch,
miss-handling, and routing policies and proposes Least-Stale eviction.
DALI (\citeproc{ref-zhu2026dali}{Zhu et al. 2026}) studies
residual-based expert prefetching in a KTransformers-based system, while
SP-MoE (\citeproc{ref-chen2025spmoe}{Chen et al. 2025}) combines expert
prefetching with speculative decoding. These works primarily ask which
mechanisms improve a specified implementation. We ask a complementary
question: what must be stated before a trace-derived policy comparison
can be interpreted or transferred?

The distinction matters for Axis I. A real execution system determines
its own transfer and retention order; our replay counterexample does not
retroactively apply to such a system. It applies to a simulator only
when the simulator claims the fused-event traffic contract of §2 while
permitting a start-resident, not-yet-served expert to be evicted and
counted again inside that event.

\subsection{12.2 Expert prediction is not next-use
prediction}\label{expert-prediction-is-not-next-use-prediction}

Fate (\citeproc{ref-fang2025fate}{Fang et al. 2025}), Pre-gated MoE
(\citeproc{ref-hwang2024pregated}{Hwang et al. 2024}), SiDA-MoE
(\citeproc{ref-du2024sidamoe}{Du et al. 2024}), and MoE-Beyond
(\citeproc{ref-gavhane2025moebeyond}{Gavhane et al. 2025}) exploit
predictability in expert activation. Their prediction targets and system
contracts differ: several predict which experts a request will select in
an upcoming layer or token so that transfer can begin early; Pre-gated
MoE changes the architecture and training procedure; MoE-Beyond predicts
token/layer expert activations from token embeddings and layer positions
and separately evaluates a trace-based cache simulation at batch size
one. Section 7 instead predicts a cached block's next-use distance for
victim ranking and bypass admission. A positive result on next-expert
prediction is therefore neither a counterexample to nor evidence for our
negative next-use result.

\subsection{12.3 Audit of reported evaluation
contracts}\label{audit-of-reported-evaluation-contracts}

We audited the public text of ten representative papers (Appendix A):
seven end-to-end offloading systems, one trace-driven cache simulation,
and two architecture/prediction studies. We record what each paper
reports, not what an unreleased implementation may do. Axis I has narrow
applicability --- for the seven end-to-end systems the implementation,
not a trace replay, defines execution, and the single trace-driven study
assumes batch size one, which removes the cross-request union central to
our conditions. Beyond that, none of the ten directly reports the
measured per-step/per-layer expert union divided by usable per-layer
capacity, so the \texttt{r\_bar} required by §6 cannot be reconstructed
without the underlying routes; and while papers identify datasets and
often prompt or generation lengths, the number of instruction templates
per category, chat-template application and positional synchronization
are generally not reported, which prevents applying the diagnostic of §5
from the paper alone. Missing fields do not imply contamination.

This audit does \textbf{not} show that any cited speedup or quality
result is wrong. Many evaluate targets outside our scope, and
real-system measurements avoid the specific replay failure by
construction. It shows that the public reports do not expose a common
evaluation contract sufficient to compare cache-policy numbers across
systems and models. Appendix A gives the per-paper evidence and puts our
own invalidated analyses in the first row.

\section{§13 Artifacts and Conclusion}\label{artifacts-and-conclusion}

\subsection{13.1 Released artifacts}\label{released-artifacts}

\textbf{Simulator.} The event-atomic replay engine of §2, with per-layer
and global cache scope, seeded tie-breaking, and the eight policies of
§2.3 including the \texttt{Belady-forced-admit} decomposition probe of
§7.2.

\textbf{Conformance trace.} The hand-verifiable trace of §2.5 --- 3
layers, 4 experts per layer, 5 scheduling steps, 30 union accesses,
yielding 12 hits and 18 misses under Definition 2 --- with the expected
counts, so that an independent implementation can confirm which
semantics it realizes.

\textbf{Workload probe set.} ControllerProbe-D1: 432 records across six
workload archetypes, twelve prompt forms crossed with two instruction
languages, eight task framings and five payload rendering styles per
archetype, a group-aware discovery/confirmatory split, and the
matched-pair fixed-template control arm of §5.5. Built entirely from
public benchmark and public industrial sources.

\textbf{Contamination audit tooling.} The prompt-side and route-side
diagnostics of §5.3--§5.4, implemented to run on any probe set and any
route collection, not only ours. This is the component we expect to be
most directly reusable.

\textbf{Routing traces and event streams.} Where upstream licences
permit redistribution, decode routing for three models, including the
corrected collection of §3.3 and the earlier single-template collection
retained as the contaminated condition. Otherwise the release contains
collection/conversion scripts, frozen selections, manifests and
cryptographic hashes rather than silently redistributing restricted
source material.

\textbf{Pre-registration documents, unchanged.} Every frozen criterion
with the split it was to be evaluated on, including those in which a
criterion failed and the associated development line was stopped, and
including the design revision recorded in §8.6.

\textbf{Derived result artifacts}, each with per-condition rows and
input manifest hashes, notably the reproducible v2 thirteen-condition
gap-invariance measurement of §7.1, the regime ablation of §6.5, the
natural-C4 false-positive check, its labelled 64-token synthetic-prefix
positive control, and the static protocol audit that freezes the
evaluated event-manifest hash, discovery pin hash, quota rule, preload
convention, and both static values used in the text.

We apply the checklist of §10 to this paper in Appendix B, including one
item we cannot complete: we set no service-rate target for the primary
model, so the bandwidth budget required by D1 is unavailable without the
hardware calibration §11 says we lack. The invalidated analyses recorded
during this work are listed in Appendix A.

\subsection{13.3 Conclusion}\label{conclusion}

We set out to measure how much room a workload-aware expert cache
controller has in MoE inference. The measurement turned out to be the
harder problem.

Three evaluation choices --- how a fused-event traffic contract is
replayed, how a workload-partitioned probe set is constructed, and what
operating regime a result sits in --- each change conclusions rather
than merely shift numbers. The first inverts policy rankings by
selectively penalizing recency-based policies. The second manufactures
apparent workload locality large enough to reverse which workloads look
cache-friendly. The third makes cross-model comparison uninterpretable
unless the ratio of per-step expert union to cache capacity is reported,
and even then does not license extrapolation. A fourth --- the service
discipline --- we found only by accident, after the first three were
already identified.

With all of them controlled, a large and stable gap to the offline
optimum remains, and it is mostly not what it appears to be:
84.3--96.6\% of it rests on knowing which resident block is used
furthest in the future, and the natural online estimator of that
quantity performs worse than the baseline it replaces. Three natural
mechanisms failed to reach pre-registered thresholds in the settings we
measured; a fourth experiment, semantic partitioning, was itself
invalidated because its dedicated partitions could not hold one
request's top-k set. A fixed, offline-chosen residency set remains a
useful transparent reference whose union-denominator traffic is
determined by an access-count distribution rather than access order. It
never wins under standard continuous batching, is close to LFRU only at
high local concurrency, and is substantially worse in the slack regimes
most relevant to our unmeasured frontier-scale motivation; it is not a
validated product mechanism.

We state our position narrowly. We have not shown that expert caching is
a closed problem, that future-victim distance is unpredictable in
principle, or that online controllers have no future. We have shown that
\textbf{in our evaluated settings a large offline-optimal gap
substantially overstates the gains recovered by representative
lightweight causal mechanisms}, and that reporting such a gap without
decomposing it invites an inference the evidence does not support.

The portable output of this work is not any individual number --- §6
establishes that the numbers are regime-bound --- but the reporting
checklist of §10, the conformance trace that pins down the replay
semantics, the contamination diagnostics that need no control arm, and
the released artifacts. We expect the checklist to be incomplete, and we
would consider a fifth confounder found by a reader to be a use of it
rather than a refutation.

\section{Declaration of generative AI and AI-assisted technologies in
the writing
process}\label{declaration-of-generative-ai-and-ai-assisted-technologies-in-the-writing-process}

During the preparation of this work the author used two generative AI
systems, in distinct roles.

Codex (OpenAI), accessed through the Codex command-line interface, was
used to implement, test and run experimental code --- the event-atomic
simulator, the workload and probe construction, the experiment
orchestration and the analysis scripts released with the paper --- and
to produce the intermediate analyses and candidate conclusions from
which the author worked.

Claude (Anthropic), accessed through the Claude Code command-line
interface, was used to draft and revise the prose of the manuscript from
the author's outlines, results and analytical decisions; to contribute
to the implementation of the same released code; to audit the
experimental claims produced in the course of the work, in an
adversarial review across the two systems in which one audited the
other's conclusions; and to check the manuscript for internal
consistency between its stated numbers and the frozen result manifests.

The author set the research questions and the standard of evidence for
the work, reviewed and adjudicated every experimental design decision
and every interpretation of results, verified each reported result
before inclusion, and is solely responsible for the claims advanced in
the paper. Neither system was used to synthesise, alter or select any
raw trace, observation or reported number: every quantitative result
here was produced by the released, versioned code from the frozen inputs
recorded in the artifact manifest, and was regenerated and checked
against that manifest. The invalidated analyses listed in Appendix A
include cases in which a model-generated conclusion was found to be
wrong and withdrawn; they are recorded because the subject of this paper
is the fragility of measurement, and assistance of this kind is one of
the ways a measurement can go wrong.

The author reviewed and edited all content and takes full responsibility
for the content of this publication.

\section{Appendix A --- Related-work reporting
audit}\label{appendix-a-related-work-reporting-audit}

Date audited: 2026-08-02.\\
Rule: \texttt{not\ reported} means only that the public paper text did
not expose the field. It is not an assertion about private code or
actual execution.

The audit freezes the following public versions: SpecMD v1, DALI v1,
Fate v2, Pre-gated MoE v3, SiDA-MoE v2, MoE-Beyond v1,
Mixtral-Offloading v1, HOBBIT v2, MoE-Infinity v3, and SP-MoE v2. Titles
and version identifiers were rechecked against the official arXiv
abstract pages on 2026-08-02; detailed evaluation claims were checked
against the corresponding PDFs. \texttt{paper/references.bib} contains
the formal records. A later paper revision must be re-audited rather
than silently substituted.

\subsection{A.1 Self-audit first}\label{a.1-self-audit-first}

{\def\LTcaptype{none} 
\begin{longtable}[]{@{}
  >{\raggedright\arraybackslash}p{(\linewidth - 6\tabcolsep) * \real{0.2500}}
  >{\raggedright\arraybackslash}p{(\linewidth - 6\tabcolsep) * \real{0.2500}}
  >{\raggedright\arraybackslash}p{(\linewidth - 6\tabcolsep) * \real{0.2500}}
  >{\raggedright\arraybackslash}p{(\linewidth - 6\tabcolsep) * \real{0.2500}}@{}}
\toprule\noalign{}
\begin{minipage}[b]{\linewidth}\raggedright
item
\end{minipage} & \begin{minipage}[b]{\linewidth}\raggedright
original state
\end{minipage} & \begin{minipage}[b]{\linewidth}\raggedright
correction
\end{minipage} & \begin{minipage}[b]{\linewidth}\raggedright
status
\end{minipage} \\
\midrule\noalign{}
\endhead
\bottomrule\noalign{}
\endlastfoot
offline oracle & forced every miss to be admitted & implemented
bypass-capable Belady; retained forced-admit only as a decomposition
probe & superseded \\
replay & flattened an event in ascending expert-ID order & event-start
hit/miss classification and boundary retention & conclusion about
admission/pinning withdrawn \\
workload probe & one category template, raw continuation, 63 decode
steps & chat template, diverse wrappers, 384-step cap, natural EOS,
staggered arrivals, matched control & industrial-locality headline
withdrawn \\
scheduling oracle & per-step best of 300 random batches: 19.3\% &
complete trajectories, \texttt{W=4}: 5.9--7.9\% & old headroom estimate
withdrawn \\
semantic partitions & best dedicated partition had 6 slots for top-8
routing & identified that the grid could not hold one request's
single-step set & treated as design failure, not mechanism refutation \\
condition count & verbal claim of 17; first 13-condition artifact lacked
frozen selections & deterministic 13-condition v2 with IDs, offsets,
condition hashes and source hash & v1 superseded \\
\end{longtable}
}

\subsection{A.2 Per-paper extraction}\label{a.2-per-paper-extraction}

Abbreviations: \textbf{A} end-to-end system; \textbf{B} trace-driven
simulation; \textbf{C} architecture/training co-design; \textbf{D}
prediction study. ``Union ratio'' means the measured per-step/per-layer
union divided by usable cache capacity, not a null model estimate.

\textbf{Table A1.} Per-paper extraction from the frozen public versions.
``NR'' means not reported in the public paper.

{\def\LTcaptype{none} 
\begin{longtable}[]{@{}
  >{\raggedright\arraybackslash}p{(\linewidth - 6\tabcolsep) * \real{0.2500}}
  >{\raggedright\arraybackslash}p{(\linewidth - 6\tabcolsep) * \real{0.2500}}
  >{\raggedright\arraybackslash}p{(\linewidth - 6\tabcolsep) * \real{0.2500}}
  >{\raggedright\arraybackslash}p{(\linewidth - 6\tabcolsep) * \real{0.2500}}@{}}
\toprule\noalign{}
\begin{minipage}[b]{\linewidth}\raggedright
work
\end{minipage} & \begin{minipage}[b]{\linewidth}\raggedright
class and evaluation vehicle
\end{minipage} & \begin{minipage}[b]{\linewidth}\raggedright
reported experimental setting
\end{minipage} & \begin{minipage}[b]{\linewidth}\raggedright
reporting interpretation
\end{minipage} \\
\midrule\noalign{}
\endhead
\bottomrule\noalign{}
\endlastfoot
\href{https://arxiv.org/abs/2602.03921}{SpecMD} & A; PyTorch hooks on
one A100-80GB with software capacity/bandwidth constraints & GSM8K,
TruthfulQA and NaturalQuestions; autoregressive generation; 1/5/25\%
capacity; 5GB/s example; active batch and chat formatting NR &
End-to-end path, so event replay is N/A; union ratio and template
multiplicity NR. Comparable policy names do not imply the same execution
contract or denominator. \\
\href{https://arxiv.org/abs/2602.03495}{DALI} & A; KTransformers-based
RTX 3090 system & C4 with WikiText calibration; default prompt/output
64; batch-size sweeps; often 50\% cache with some 25\% ablations; chat
formatting and template multiplicity NR & Event replay is N/A and union
ratio is NR. DALI reports real throughput; Axis I is not a criticism of
it. \\
\href{https://arxiv.org/abs/2502.12224}{Fate} & A/D; two-PC
implementation & ChatGPT-prompts, HumanEval and GSM8K; decode output up
to 1024; prefill lengths 128/256/512; memory budgets reported; active
decode concurrency, chat formatting and template multiplicity NR & Event
replay is N/A and union ratio is NR. Fate predicts next experts for
prefetch, not next-use distance. \\
\href{https://arxiv.org/abs/2308.12066}{Pre-gated MoE} & C; trained
architecture and FasterTransformer implementation & XSum, closed-book QA
and SQuAD; primary system focus at batch 1; model/cache design and task
datasets reported; chat template not applicable or NR & Event replay is
not the paper's question and union ratio is NR. The method modifies
training and the routing contract, outside strictly post-hoc caching. \\
\href{https://arxiv.org/abs/2310.18859}{SiDA-MoE} & C/D; hash-based
predictor/system design & C4 and Switch-family evaluation; model
configuration reported; category-template risk not applicable to the
reported corpus setup & Event replay is not the primary question and
union ratio is NR. Its prediction objective differs from victim next-use
ranking. \\
\href{https://arxiv.org/abs/2508.17137v1}{MoE-Beyond v1} & B/D;
token/layer activation predictor plus trace-based simulator & 6,994
Puffin training prompts and 100 WebGLM-QA test prompts; batch size 1;
cache-capacity sweep; chat formatting NR & Token-by-token replay is
described, but fused-event retention details and union ratio are NR.
Batch 1 removes cross-request union, leaving insufficient detail for an
event-contract comparison. \\
\href{https://arxiv.org/abs/2312.17238}{Mixtral-Offloading} & A;
end-to-end offloading implementation & OpenAssistant conversations with
autoregressive sampling; practical focus at batch 1; per-layer cache
sizes and source conversations reported; chat-template details NR &
Event replay is N/A and union ratio is NR. Its real-system latency is
not invalidated by our replay result. \\
\href{https://arxiv.org/abs/2411.01433}{HOBBIT} & A; llama.cpp-based
CPU--GPU system & Alpaca for performance; GSM8K and TruthfulQA for
quality; several input/output lengths; batch 1; cache/precision budgets
reported; chat formatting and template multiplicity NR & Event replay is
N/A and union ratio is NR. Lossy precision fallback is outside our
strictly lossless scope. \\
\href{https://arxiv.org/abs/2401.14361}{MoE-Infinity} & A; RTX A5000
end-to-end system & 290 BIGBench/FLAN/MMLU tasks; prompt 512/output 32
in the main latency test; local deployment, effectively batch 1; task
rendering insufficient for §5 diagnostics & Event replay is N/A and
union ratio is NR. Request-level activation maps and real execution
answer a different question. \\
\href{https://arxiv.org/abs/2510.10302}{SP-MoE} & A; speculative
decoding and expert-prefetch system & Four public datasets;
autoregressive evaluation; batch size 1; chat/template multiplicity NR &
Event replay is N/A and union ratio is NR. Its prefetch/speculation
target differs from cache-victim prediction. \\
\end{longtable}
}

\subsection{A.3 What may and may not be
concluded}\label{a.3-what-may-and-may-not-be-concluded}

The audit supports three reporting claims: most audited systems do not
expose a directly comparable \texttt{r\_bar}; prompt-surface
contamination cannot generally be checked from their papers; and our
replay-semantics warning should not be aimed at end-to-end
implementations. It does not support re-labelling published speedups as
artifacts, assigning a direction to an unreported implementation choice,
or inferring that a dataset used one fixed prompt merely because prompt
construction was omitted.

\section{Appendix B --- Supplementary
material}\label{appendix-b-supplementary-material}

Material referenced from the body but not required to follow the
argument.

\subsection{B.1 A minimal reporting
block}\label{b.1-a-minimal-reporting-block}

The following is compact enough to include verbatim in a paper or
artifact README, and covers the items above that admit a one-line
answer.

\begin{verbatim}
Execution model
  event unit ......................... (step, layer) union | per-access
  retention decision ................. event boundary | immediate
  intra-event order (if immediate) ... ascending id | randomized | n/a
  conformance trace .................. yes | no

Workload
  templates per category ............. N
  distinct prompt heads / finals ..... N / N
  chat template applied .............. yes | no
  decode steps recorded .............. N
  position-locked cohorts ............ yes | no
  unique generated sequences ......... N / N
  positional token agreement ......... mean, p95
  expert Jaccard | low-text-overlap .. value (cross-category baseline: value)
  effect stable across position bands  yes | no | n/a

Regime
  experts N, top-k, layers ........... N / k / L
  residency fraction rho ............. value
  cache scope ........................ per-layer | global
  union per layer-step ............... mean, p95
  capacity per layer ................. value
  r(s,l) = union / capacity .......... mean, p95, fraction > 1
  concurrency ........................ B (and admitted sessions if rotating)
  service discipline ................. continuous | rotation(W=..) | other
  miss denominator ................... pre-dedup | post-dedup
  dedup credited to policy ........... yes | no
  static fit source .................. same trace | discovery split | n/a
  static quota rule .................. description | n/a
  initial pinned load counted ........ yes | no | n/a

Oracle
  offline bound reported ............. yes | no
  forced-admission decomposition ..... admission % / future-victim %
  absolute bytes per output token .... value
  bandwidth budget assumed ........... value (and target service rate)
\end{verbatim}

\subsection{B.2 The checklist applied to this
paper}\label{b.2-the-checklist-applied-to-this-paper}

We fill the reporting block of §10.5 for the primary configuration. One
field we cannot complete, and we say so rather than omitting it.

\begin{verbatim}
Execution model
  event unit ......................... (step, layer) union
  retention decision ................. event boundary
  intra-event order .................. n/a
  conformance trace .................. yes (§2.5)

Workload
  templates per category ............. 24 form x language (diverse arm)
                                       1 (matched control arm, by design)
  distinct prompt heads / finals ..... 23-24 / 16-22 per 24-record cell
  chat template applied .............. yes, reasoning mode disabled
  decode steps recorded .............. up to 384, natural EOS
  position-locked cohorts ............ no (arrival offsets 0-23)
  unique generated sequences ......... 12 / 12 per archetype sample
  positional token agreement ......... 0.45-1.06% mean
  expert Jaccard | low-text-overlap .. 0.085-0.144 (cross-category: 0.077)
  effect stable across position bands  yes for 2 of 6 archetypes (§5.4, §5.7)

Regime
  experts N, top-k, layers ........... 128 / 8 / 48
  residency fraction rho ............. 0.40
  cache scope ........................ per-layer, remainder to lowest indices
  union per layer-step ............... 42.63 mean, 50 p95
  capacity per layer ................. 51-52
  r(s,l) = union / capacity .......... mean 0.833; fraction >1: 2.19%
  concurrency ........................ B = 8
  service discipline ................. continuous batching (FCFS);
                                       rotation variant reported separately
  miss denominator ................... pre-dedup (logical assignments)
  dedup credited to policy ........... no
  static fit source .................. discovery split (Static-frozen);
                                       evaluation trace (diagnostic only)
  static quota rule .................. equal per layer; remainder to lowest indices
  initial pinned load counted ........ yes

Oracle
  offline bound reported ............. yes
  forced-admission decomposition ..... 15.69% / 84.31%  (B=8)
                                       3.40% / 96.60%   (B=2)
  absolute transfer .................. 69.2 blocks per output token
                                       (best causal, held-out)
  bandwidth budget assumed ........... NOT APPLICABLE - no service-rate target
                                       is assumed for the primary model
\end{verbatim}

The last field is a gap in our own reporting. Item D1 asks that a gap be
accompanied by the bandwidth budget implied by a target service rate,
and for the primary model we set no such target: the measurements are in
blocks, and converting them to a budget would require the hardware
calibration §11.1 says we do not have. We record the block count
instead, and note that a reader wishing to apply D1 to our numbers must
supply the missing conversion themselves.

For completeness, the failures recorded during this work and released
with the pre-registrations are: an initial offline bound that forced
admission of every miss and was therefore not the offline optimum; the
sequential replay of §4, whose correction withdrew a conclusion about
admission and pinning; the single-template probe set of §5, whose
correction withdrew a confirmatory conclusion about industrial workload
locality; a scheduling headroom estimate of 19.3\% that fell to
5.9--7.9\% when measured on complete trajectories (§8.4); a category
partitioning experiment that could not have produced a positive result
(§8.2); and, during internal review, an unsourced condition count that
was corrected from seventeen to thirteen. The first 13-condition
artifact still lacked frozen request selections and was superseded by a
deterministic v2 with every request ID, arrival offset, condition hash
and source-manifest hash.

\subsection{B.3 Does the static disadvantage grow with expert
count?}\label{b.3-does-the-static-disadvantage-grow-with-expert-count}

Referenced from §9. \textbf{Hypothesis-generating only}: two comparable
points on the expert-count axis, heterogeneous models, and all four rows
collected under the single-template protocol of §5, whose surface
repetition inflates temporal locality and therefore plausibly inflates
the static disadvantage in every row. The comparison is internally
consistent; the levels are not trustworthy, and a two-point
extrapolation across a further factor of seven in expert count would not
be.

Table 18 measures one model. Comparing across models requires aligning
on the regime variable of §6, and doing so suggests a trend we can
report but not establish.

\textbf{Table B1.} Static versus dynamic at aligned operating regimes.
\emph{(\ensuremath{\rho} = 40\%, per-layer scope. Pin lists fitted on
the evaluation trace itself --- deliberately leaky, so the static arm is
shown at its best case and the trend is conservative.)}

{\def\LTcaptype{none} 
\begin{longtable}[]{@{}lrrrrr@{}}
\toprule\noalign{}
model & experts & \texttt{r\_bar} & static & LFRU & static vs LFRU \\
\midrule\noalign{}
\endhead
\bottomrule\noalign{}
\endlastfoot
40-expert & 40 & 0.500 & 31.15\% & 25.91\% & \textminus{}20.2\% \\
64-expert & 64 & 0.312 & 35.64\% & 25.65\% &
\textbf{\textminus{}39.0\%} \\
128-expert & 128 & 0.293 & 19.57\% & 12.83\% &
\textbf{\textminus{}52.6\%} \\
128-expert & 128 & 0.156 & 19.87\% & 10.57\% & \textminus{}87.9\% \\
\end{longtable}
}

Two directions are visible: at comparable \texttt{r\_bar}, the
disadvantage grows with expert count (\textminus{}39.0\% at 64 experts,
\textminus{}52.6\% at 128); and within a single model it grows as the
regime becomes slacker (\textminus{}52.6\% at \texttt{r\_bar} = 0.29,
\textminus{}87.9\% at \texttt{r\_bar} = 0.16). Both are consistent with
the mechanism of §9.1 --- more experts means a flatter popularity
distribution, so a fixed top-fraction captures less of the traffic,
while a dynamic policy still tracks a working set that remains small
relative to capacity.

We label this hypothesis-generating rather than a result. There are only
two comparable points on the expert-count axis; the models differ in
training and tokenizer as well as in expert count; and all four rows use
the single-template collection of §5, whose surface repetition inflates
temporal locality and therefore plausibly inflates the static
disadvantage in every row. The comparison is internally consistent, but
the levels are not trustworthy and a two-point extrapolation across a
further factor of seven in expert count would not be.

\subsection{B.4 The invalidated partition
grid}\label{b.4-the-invalidated-partition-grid}

Referenced from §8.2. Retained so that the design failure is
inspectable, not as a negative result for semantic partitioning.

The partitioning experiment gave a dedicated per-layer slot budget to
each of the two archetypes that §5.7 identified as genuinely
cache-friendly, with the remaining four sharing a third partition. Total
capacity was held exactly equal to the shared baseline and the remainder
was assigned to the shared partition, so the dedicated partitions
received no arithmetic advantage. A seven-point allocation grid was
swept on the calibration split, the best allocation frozen, and the
held-out split evaluated once. Every allocation lost, and the frozen one
increased transfer by 115.9\%.

The natural reading is that semantic partitioning destroys
cross-category expert sharing and duplicates commonly used experts
across partitions. We tested that reading and it is largely wrong.
Running the \emph{shared} cache at each partition's own slot count gives
61.08\% miss at 6 slots and 27.25\% at 39 slots, against 18.76\% at the
full 51. Weighting these by each partition's share of logical expert
assignments (0.1415, 0.1557, 0.7028) predicts

\begin{quote}
0.1415 \ensuremath{\times} 61.08\% + 0.1557 \ensuremath{\times} 61.08\%
+ 0.7028 \ensuremath{\times} 27.25\% = \textbf{37.30\%}
\end{quote}

against a measured 37.99\%. \textbf{Capacity fragmentation alone
accounts for 96\% of the degradation}; sharing and duplication together
account for the remaining 4\%.

The root cause is arithmetic. Top-\texttt{k} is 8, so a single request
touches eight experts at every layer, while a dedicated partition in the
best allocation holds six. No partition in the grid could retain even
one request's per-step working set, and the grid's largest dedicated
allocation, 16 slots, holds two. \textbf{The experiment as designed
could not have produced a positive result}, and we say so rather than
presenting the outcome as a mechanism-level refutation.

Whether the arithmetic generalizes is a separate question, and for the
frontier-scale reference frame of §3.2 it does not: 358 per-layer slots
divided six ways gives approximately 59.7 per partition against a
per-category per-step union of roughly 16--26, a margin of
2.3--3.7\ensuremath{\times}. Hard partitioning may still lose there for
the sharing and duplication reasons, but it would not be excluded by
capacity arithmetic, and the \textminus{}115.9\% figure must not be
carried across.

\section*{References}\label{bibliography}
\addcontentsline{toc}{section}{References}

\protect\phantomsection\label{refs}

\end{document}